\documentclass[11pt]{article}

\usepackage[final]{acl}

\usepackage{times}
\usepackage{latexsym}

\usepackage[T1]{fontenc}

\usepackage[utf8]{inputenc}

\usepackage{microtype}

\usepackage{inconsolata}

\usepackage{graphicx}

\usepackage{amsfonts}   
\usepackage{xspace}     
\providecommand{\texorpdfstring}[2]{#1}
\newcommand{\df}{\texorpdfstring{$\mathbb{DF}$}{DF}\xspace}
\newcommand{\pf}{\texorpdfstring{$\mathbb{PF}$}{PF}\xspace}

\newcommand{\lap}{\textit{LLM-as-planner}\xspace}
\newcommand{\laf}{\textit{LLM-as-formalizer}\xspace}
\newcommand{\method}{\texorpdfstring{\texttt{PDDL\allowbreak ego+}}{PDDLego+}\xspace}

\newcommand{\othree}{\texttt{o3-mini}\xspace}
\usepackage{amsmath}
\usepackage{booktabs}    
\usepackage{listings}    
\usepackage{placeins}    
\usepackage{enumitem}
\setlist[enumerate]{topsep=0pt, partopsep=0pt, parsep=0pt, itemsep=0pt, leftmargin=*}
\setlist[itemize]{topsep=0pt, partopsep=0pt, parsep=0pt, itemsep=0pt, leftmargin=*}

\title{Iterative Formalization and Planning in Partially Observable Environments}

\author{
\textbf{Liancheng Gong}$^{1}$ \quad
\textbf{Wang Zhu}$^{2}$ \quad
\textbf{Jesse Thomason}$^{2}$ \quad
\textbf{Li Zhang}$^{1}$ \\
$^{1}$Drexel University \quad 
$^{2}$University of Southern California \\
\texttt{gonglc@umd.edu},
\texttt{harry.zhang@drexel.edu}
}

\begin{document}
\maketitle
\begin{abstract}
Using LLMs not to predict plans but to formalize an environment into the Planning Domain Definition Language (PDDL) has been shown to improve performance and control. While most existing methodology only applies to fully observable environments, we adapt to the more realistic and challenging partially observable environments without sufficient information to make a complete plan. We propose \method, a framework to iteratively formalize, plan, grow, and refine PDDL representations by decomposing the environment and the goal into fully observable episodes. Without fine‑tuning, in-context exemplars, or trajectories, \method improves planning success and exhibits robustness against problem complexity compared to end-to-end approaches. We also show that the domain knowledge captured after a successful trial can benefit future tasks. \footnote{Our resources can be found at \url{https://github.com/zharry29/pddlego-plus}.}
\end{abstract}

\section{Introduction}

Whether large language models (LLMs) can plan given a grounded environment has been at the center of attention of the AI planning community \cite{huang2025limitlanguagemodelsplanning,hu2025text2worldbenchmarkinglargelanguage,ishay2025llmalbridginglargelanguage,parmar2025plangenmultiagentframeworkgenerating,hao2025planningrigorgeneralpurposezeroshot}. Most efforts that use LLMs for planning involve one of two methodologies. The first, \lap, directly generates a plan \cite{parmar2025plangenmultiagentframeworkgenerating,stein2025automatinggenerationpromptsllmbased,valmeekam2024planbench}. The second, \laf, instead generates a representation of the environment and the task that can be solved by a formal planner \cite{hu2025text2worldbenchmarkinglargelanguage,zuo2024planetarium,xie2023translating}. We focus on \laf and consider \lap as a baseline due to the former's good performance, interpretability, control, and formal guarantee, shown in existing work. 

\begin{figure}[t!]
    \centering
    \includegraphics[width=\columnwidth]{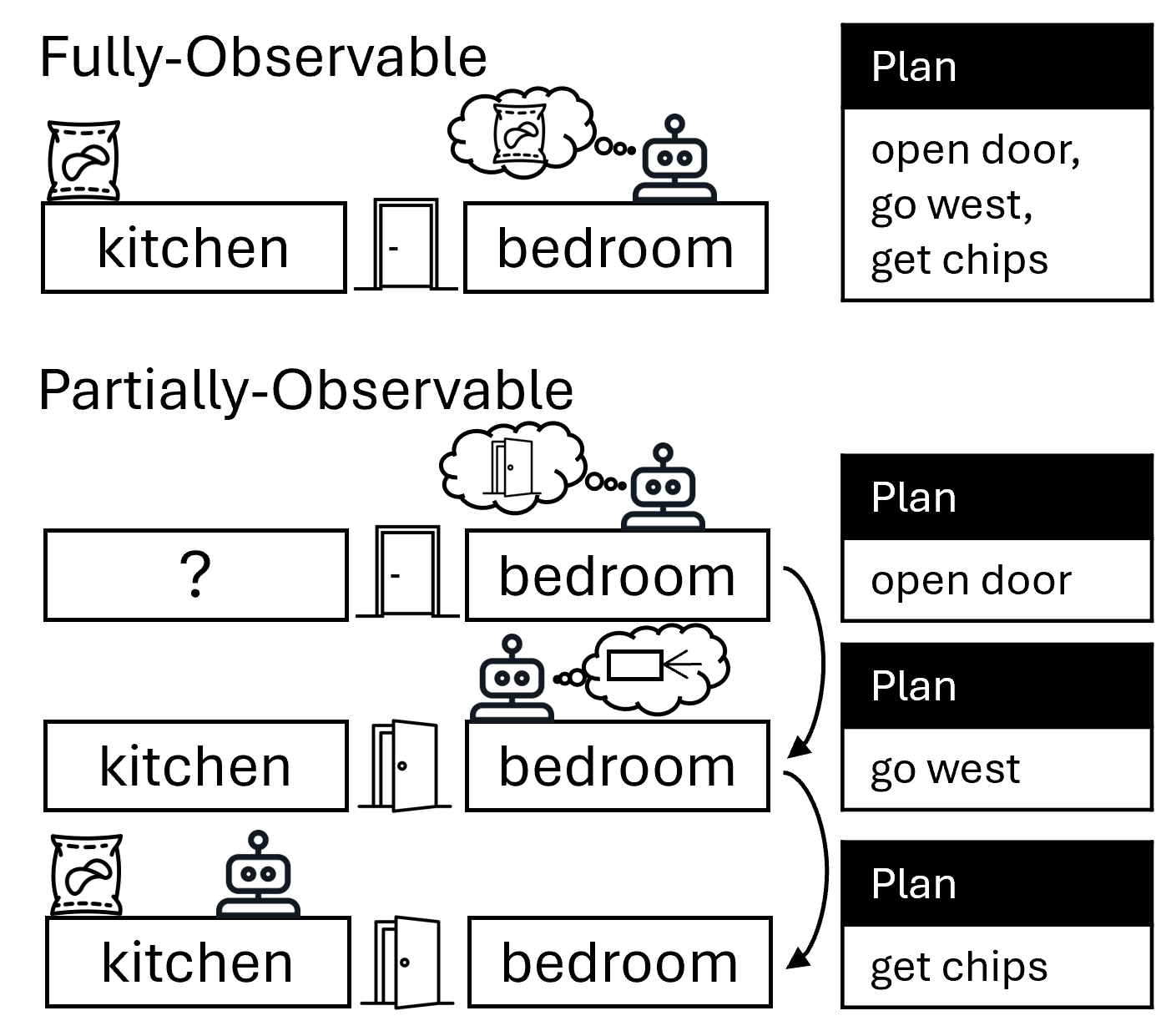}
    \caption{Unlike fully-observable environments (upper), partially observable environments (lower) require planning agents to make partial plans, interact with the environment, and gain new observations.}
    \label{fig:pomdp}
\end{figure}

Most work studied \textit{fully observable environments} \cite{li2024embodied,zhu2024language,liu2023llm+}, where all initial and eventual entity states are known throughout the planning task (e.g., a robot fully aware of the layout of a house retrieves an object from a known location). Methodology proposed in this setting cannot be readily applied to the more realistic and challenging \textit{partially observable environments}. Here, the model has only access to a subset of the environment, unable to propose a complete plan until sufficient exploration (e.g., a robot is initially deployed in a house with obstructed views to look for an object). Only a handful of existing studies \cite{tang2024worldcodermodelbasedllmagent,zhang-etal-2024-pddlego,wong2023learning} have tackled this challenge but suffer from three shortcomings. First, some assumed certain components of the planning representation such as predicates, action semantics, or entity states, thus simplifying the task. Second, some only considered simplistic, one-attempt formalization technique, underestimating the potential of the method. Third, some rely on existing trajectories as in-context exemplars, which are not readily accessible in real-life applications. 

\begin{figure}[t!]
    \centering
    \includegraphics[width=\columnwidth]{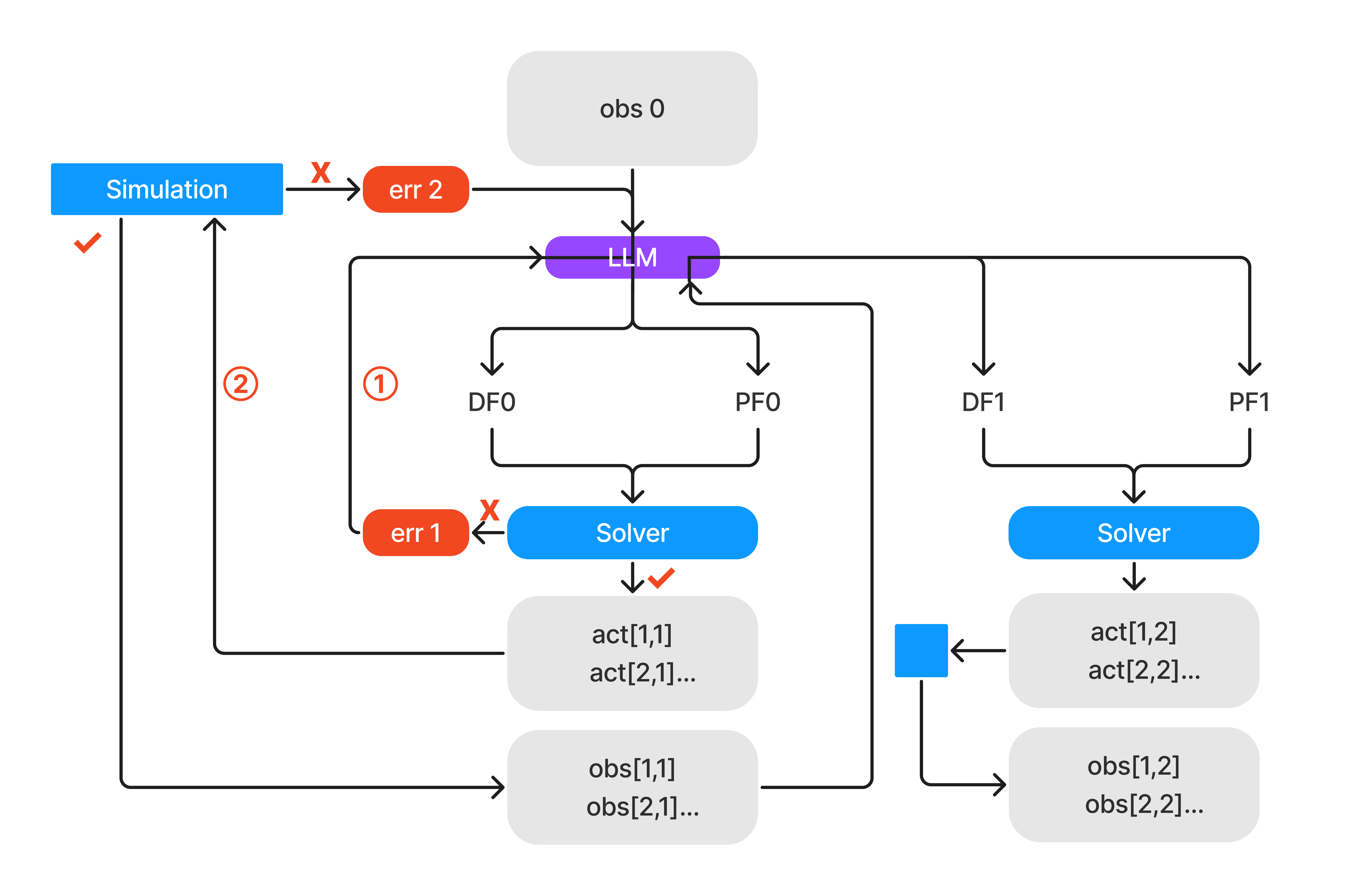}
    \caption{An illustration of \method, using \laf. Input environmental observations into LLM to generate PDDL representations, which are input into solver to output an action plan. The plan is executed in simulation resulting in new observations to grow PDDL. When errors occur, the LLM refines the PDDL. Unlike PDDLego, which assumes a fixed domain file, \method revises both DF and PF throughout interaction.}
    \label{fig:pddlegoplus}
\end{figure}

We address all those shortcomings by proposing \method (Figure~\ref{fig:pddlegoplus}), a framework to iteratively formalize, plan, grow, and refine the representation of a textual simulated environment. \method operates without fine-tuning, in-context exemplars, access to any prior trajectories, or any non-essential components of the target representation. We assume the agent is exposed to the environment for the first time, only prompted with critical and idiosyncratic information about the environment. Given the initial observation, our LLM generates full domain and problem definitions in the Planning Domain Definition Language (PDDL). Due to the partial observability, the LLM cannot initially define sufficient facts that lead to a complete plan towards the goal. Instead, we have the LLM define known facts that constitute a fully observable sub-environment and predict a feasible sub-goal. This PDDL representation is then input into a formal solver that outputs a plan of actions to be executed. If successful, the LLM iteratively grows the representation based on the new observations. If failed, we build on existing methodology of fully observable PDDL generation that revises the representation based on solver error as feedback. As interaction with the environment is possible, we design a two-stage, hierarchical refinement process where the LLM refines the representation based on both solver and simulation errors. 

We base our evaluation on two common partially observable simulated environments: CoinCollector \cite{yuan2019counting}, which is a navigation task; and ALFWorld \cite{ALFWorld20}, which is an object placement task. \method using \othree achieves 86\% and 38\% success rate, greatly outperforming the \lap baseline (52\% and 5\%) and \citet{zhang-etal-2024-pddlego} (49\% and 3\%) without refinement by error. Our comprehensive analysis shows that \method is interpretable, robust to complexity, and its learned domain knowledge can be effectively reused for future tasks. 

\section{Related Work}

\textbf{Planning with LLMs} has attracted much attention. Some use LLMs for informal planning, also known as script or procedure learning \cite{zhang-etal-2020-analogous,lyu-etal-2021-goal,lal-etal-2024-tailoring}. However, even the most coherent and plausible informal plans lack executability and verifiability.  In LLM formal planning realm, work can be divided into four quadrants by two axes: whether the methodology is \lap or \laf, and whether the environment is fully or partially observable. 

Most existing research targeted \textbf{fully observable environments} for their simplicity and ease for evaluation, using \lap \cite{valmeekam2023planningabilitieslargelanguage,kambhampati2024llms,valmeekam2024planbench,stechly2024chain,stein2025automatinggenerationpromptsllmbased} and \laf \cite{hu2025text2worldbenchmarkinglargelanguage,ishay2025llmalbridginglargelanguage,parmar2025plangenmultiagentframeworkgenerating,hao2025planningrigorgeneralpurposezeroshot,zuo2024planetarium,xie2023translating,liu2023llm+,guan2023leveraging,lyu-etal-2023-faithful}. The former group mostly argued \lap's limitation without specialized techniques due to LLMs' lack of reasoning ability. The latter advocated for \laf's effectiveness due to LLMs' strong ability of structured generation, not to mention the improved interpretability and more formal guarantee. \citet{huang2025limitlanguagemodelsplanning} closely compared the two methodologies and suggested that both have merits depending on the nature of the LLMs, especially the latest ones trained to generate reasoning tokens such as \texttt{DeepSeek-R1} or o-series OpenAI models. Regardless, all of the above work on fully observable environments makes a closed‑world assumption: the initial state, goal, and truth value of every predicate are fully known at compilation time. Any methodology based on this assumption breaks down in partially observable settings because the agent sees only fragments of the world and cannot enumerate all objects or relations up‑front. 

In contrast, \textbf{partially observable environments}, despite their common occurrence in applications like robotics or web agents, has been less explored likely due to their complexity. The majority of existing work considered \lap \cite{majumder2023clin,lin2023swiftsage}, while only a few considered \laf \cite{tang2024worldcodermodelbasedllmagent,wong2023learning} like us. The most similar work to ours is \citet{zhang-etal-2024-pddlego} which proposed a preliminary methodology to iteratively formalize PDDL. Yet, we improve in two key respects.

\textbf{(i) Task‑wise novelty (full domain inference).} While they assumed an existing domain file, we infer \emph{both} files from scratch.  This difference is substantial: synthesizing the domain file is hard, requiring discovery and abstraction of types, predicates, and action semantics, while synthesizing the problem file is easier, requiring only translating observations into formalism. Analogous to program synthesis, domain is akin to classes and functions while problem is akin to function calls or unit tests. 

\textbf{(ii) Method‑wise novelty (two‑phase error refinement).} While they did not have mechanism to refine PDDL, we introduce a nested refinement loop that first localizes immediate, syntactic solver errors and then the more distant, semantic simulation errors. The two-level error signals are unique challenges that accompanies formal planning. We will describe our error routing in next section.

\paragraph{Distinction from PDDLego.}
PDDLego+ differs from \citet{zhang-etal-2024-pddlego} by (i) inferring \emph{both} the domain and problem files (DF+PF) rather than assuming a provided DF, and (ii) introducing a two-phase refinement loop that separately routes solver vs.\ simulation errors.
This refinement is not a generic self-reflection heuristic; it is induced by the solver--simulator separation in formal planning.

\noindent \textbf{Code generation with LLMs} \cite{chen2021evaluating} is at the core of our methodology and of any work on \laf. In addition to the common usage of writing or debugging programs \cite{jiang2024survey}, LLMs can also generate formal representation. For planning, while PDDL is the most popular choice in literature other work considers other languages \cite{gao2023pal,wang2023bytesized32corpuschallengetask,tang2024worldcodermodelbasedllmagent,ishay2025llmalbridginglargelanguage,hao2025planningrigorgeneralpurposezeroshot}. Practically, the choice of the planning language depends on both its expressivity and the ease of LLM generation.
Because any declarative planning language needs to conform to a certain syntax, it is low-resource and challenging to generate for LLMs \cite{ahmed2024studyingllmperformanceclosed,Blinn_2024,cassano2022multiplescalableextensibleapproach}. Indeed, \citet{huang2025limitlanguagemodelsplanning} showed that smaller LLMs struggle to generate syntactically correct PDDL, which concurs with our preliminary experiments.

\begin{figure}[t!]
    \centering
    \includegraphics[width=\columnwidth]{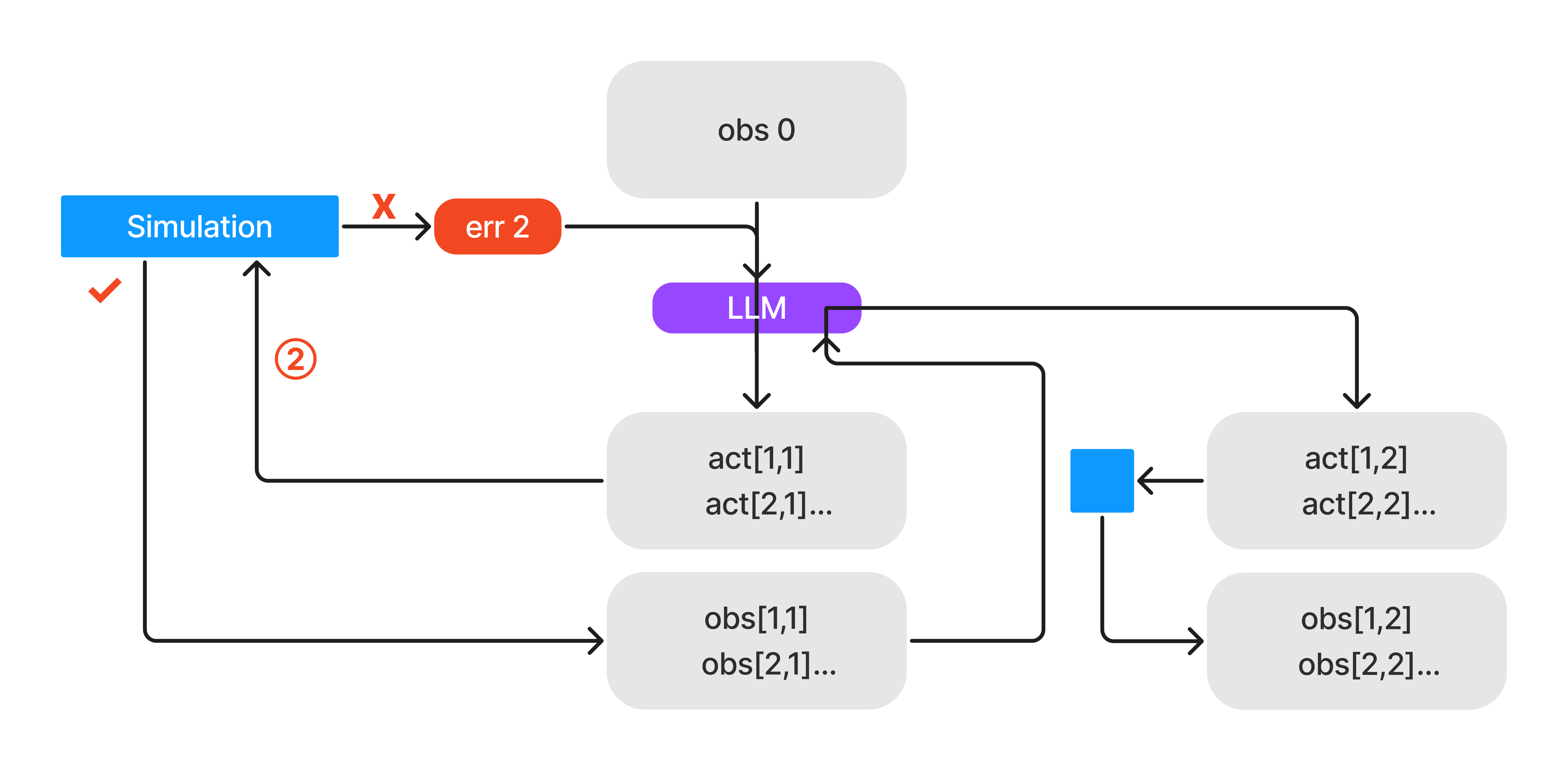}
    \caption{An illustration of a framework based on \lap which we consider as a baseline. The LLM directly generates an action plan to be executed.}
    \label{fig:baseline}
\end{figure}

\section{Methodology}

\paragraph{Partial Observable Environments.}
We work with textual, interactive, simulated environments that can be modeled as a partially observable Markov decision process. At each step, the agent receives only partial textual observations (e.g., a description of the location the agent is at) rather than a complete state (e.g., descriptions of all locations). Therefore, the agent cannot create a complete plan until it has sufficiently explored the environment to uncover missing information. Planning in partially observable environments is especially challenging for \laf due to the completeness assumption of planning languages like PDDL, requiring techniques such as goal decomposition and iterative generation \cite{zhang-etal-2024-pddlego}.

\paragraph{Proposed Method.}
Our \method framework is illustrated in Figure~\ref{fig:pddlegoplus}. To summarize, \method iteratively generates and refines a PDDL domain file (\df, defining types, predicates, and actions) and a problem file (\pf, defining objects, initial states, and goal states) by interacting with the environment. Unlike most existing work on \laf that only predicted a portion of the \df or the \pf, we predict a complete PDDL representation while minimally assuming the names and parameter lists of the actions. At each step, an LLM takes as input an observation (e.g., ``You are in the kitchen. To the South you see the corridor.'') and outputs the PDDL, which is then input into a formal solver that searches for a plan (e.g., ``move south''). The plan consisting of a sequence of actions is executed in the simulated environment. If successful, the simulation provides a new observation (e.g., ``You are in the corridor. To the North you see the kitchen.''), which is again input into the LLM to generate an \textit{updated} PDDL. 

In this process, potential errors may occur from two main sources. First, if the solver fails to find a plan (e.g., the generated PDDL has syntax errors or defines an unachievable), a \textbf{solver error} arises. Second, if the solver does find a plan but the plan cannot be fully executed in the simulated environment (e.g., trying to ``move east'' without first performing ``open door to east''), a \textbf{simulation error} arises. In either case, error messages will be input back to the LLM to generate a \textit{refined} PDDL. 

Our error refinement mechanism requires informative failure messages (e.g., “You cannot go there; door is closed’’) instead of uninformative ones (e.g., ``nothing happened’’). Similarly, we provide some general information of each environment to ensure a fair evaluation (see details in Appendix E).

\paragraph{Formal Definition.}

In every trial, an environment in the initial state of $S_0$ is instantiated and a goal state $S_N$ is provided in natural language. Each state $S_i$ at time step $i$ yields an observation $\mathrm{obs}_{i}$ in natural language alongside general information of the environment. This is used to generate the \df $df_i$ and \pf $pf_i$ initially, and \textit{update} them afterwards:
\begin{equation} \label{eq:generate}
\begin{split}
\mathrm{df}_0,\mathrm{pf}_0&=\text{LLM}(\mathrm{obs}_0) \\
\mathrm{df}_{i+1},\mathrm{pf}_{i+1}&=\text{LLM}(\mathrm{obs}_{i+1},df_i,pf_i)
\end{split}
\end{equation}
The PDDL files are used by the solver to find a plan to execute in the simulated environment:
\begin{equation} \label{eq:solve}
\begin{split}
    p_i&= [a_i^0, \ldots , a_i^{J}] \\
    &=\text{solver}(df_{i},pf_{i})
\end{split}
\end{equation}
At time step $i$, an agent proposes a plan $p_i$ of ${J+1}$ actions $a_i^j$. Every successfully executed action modifies the state within time step $i$:
\begin{equation} \label{eq:simulate_i}
S_{i}^{j+1}, \mathrm{obs}_{i}^{j+1}=\text{simulation}(a_i^j, S_i^j)
\end{equation}

All observations $\mathrm{obs}_{i}^{j}$ from plan $p_i$ are concatenated as the new observation ${obs}_{i+1}$ used in Eq.~\ref{eq:generate}:
\begin{equation} \label{eq:observation} 
    \mathrm{obs}_{i+1}= [\mathrm{obs}_{i}^{1}, \ldots , \mathrm{obs}_{i}^{J+1}]
\end{equation}
Every successfully executed plan progresses the time step by $1$:
\begin{equation} \label{eq:simulate}
\begin{split}
    S_{i+1}, \mathrm{obs}_{i}^{J+1}&=\text{simulation}(p_i,S_i) \\
    &= \text{simulation}(a_i^J, S_i^J)
\end{split}
\end{equation}

Therefore, a successful agent will find a sequence of $N$ plans that transitions the state of the environment from the initial state to the goal state:
\begin{equation}
S_N, \mathrm{obs}_{N-1}^{J+1} = \text{simulation}(p_{N-1},S_{N-1})
\end{equation}
In summary, the LLM uses the observation to generate PDDL (Eq.~\ref{eq:generate}), the solver uses PDDL to find a plan (Eq.~\ref{eq:solve}), and the simulation executes the plan to provide new observations (Eq.~\ref{eq:simulate}). 

Two errors can occur. When a \textbf{solver errors} $\mathrm{err}_\text{sol}$ appears, the error message is fed back into the LLM to \textit{refine} the \df and \pf.  When a \textbf{simulation error} $\mathrm{err}_\text{sim}$ occurs for any action in the plan, all actions already executed are nullified, the environmental state is reset, and the error message is also fed back into the LLM to \textit{refine} the \df and \pf. 

In other words, the refinement based on solver errors is an inner loop (\textcircled{1} in Figure~\ref{fig:pddlegoplus}), while that based on simulation errors is an outer loop (\textcircled{2}).

At the \(i\)-th time step, \(j\)-th outer loop iteration, \(k\)-th inner loop iteration, the LLM yields: 
\[\mathrm{df}_{i}^{j+1},\mathrm{pf}_{i}^{j+1}=\text{LLM}(\mathrm{err}_\text{sim}, \mathrm{df}_{i}^{j},\mathrm{pf}_{i}^{j})\]
\[\mathrm{df}_{i}^{j,k+1},\mathrm{pf}_{i}^{j,k+1}=\text{LLM}(\mathrm{err}_\text{sol}, \mathrm{df}_{i}^{j,k},\mathrm{pf}_{i}^{j,k})\]
In case of neither of the errors, we progress to the next time step.

In Figure~\ref{fig:pddlegoplus}, the \textit{updating} of the PDDL representation with the progression of time step (in case of no errors) is shown horizontally. The \textit{refinement} of the PDDL representation within a time step in case of errors is shown vertically. An example step \(i\) might contain the following: 
\begin{enumerate}
\item Generate \(\mathrm{df}_{i}=\mathrm{df}_{i}^{0,0}\) and \(\mathrm{pf}_{i}=\mathrm{pf}_{i}^{0,0}\). 
\item Enter the \textbf{inner loop} to refine \(\mathrm{df}_{i}^{j,k}\) and \(\mathrm{pf}_{i}^{j,k}\) until the solver outputs a valid plan or a retry limit is reached. 
\item Enter the \textbf{outer loop} to refine \(\mathrm{df}_{i}^{j}\) and \(\mathrm{pf}_{i}^{j}\) until the simulation successfully executes the plan or a retry limit is reached.
\end{enumerate} 
In this way, \method not only iteratively refines generated \df and \pf based on error messages, but also updates them based on newly uncovered information in the partially observable environment.

\begin{figure}[t!]
    \centering
    \includegraphics[width=\columnwidth]{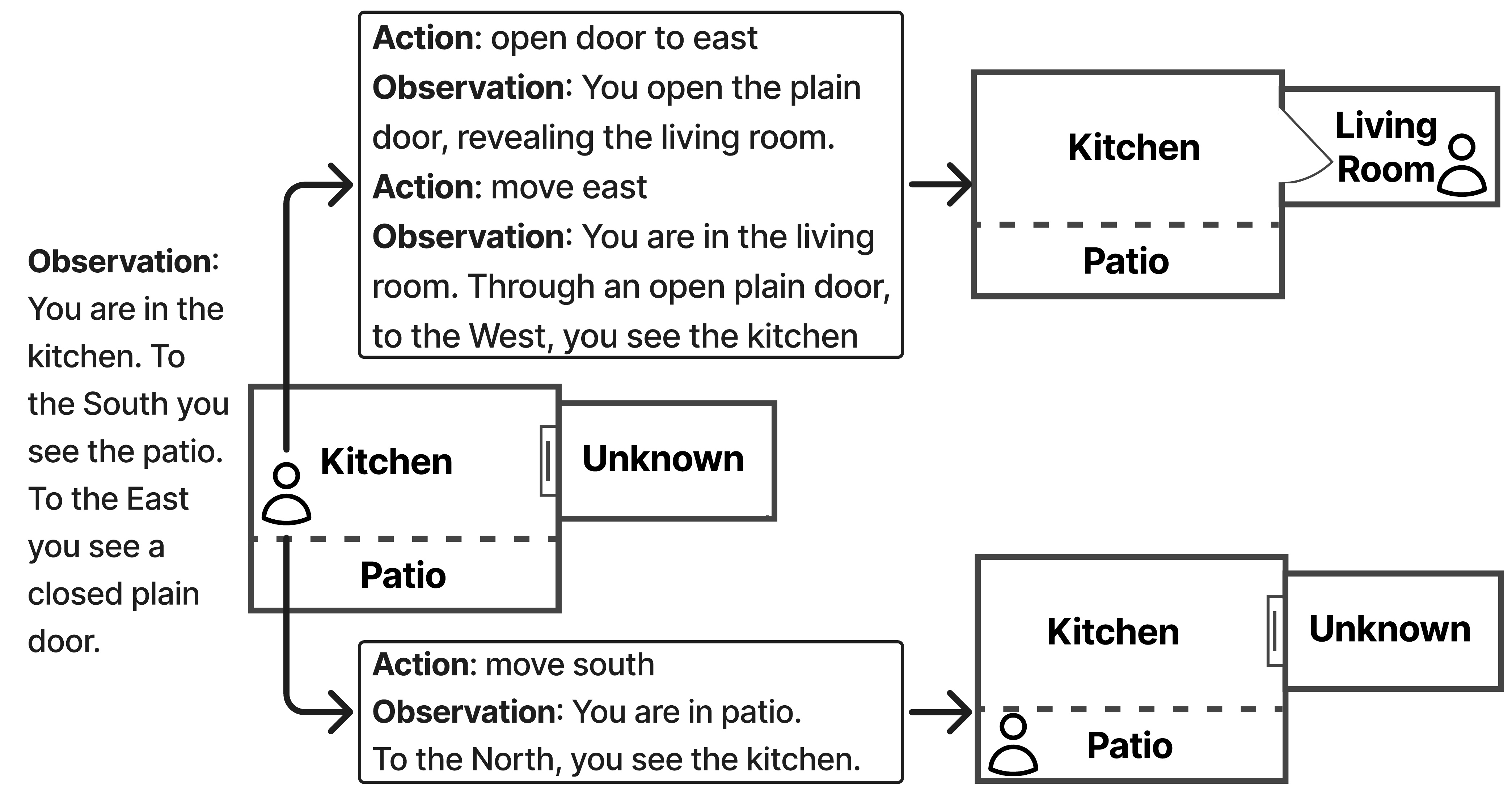}
    \caption{Illustration of the CoinCollector environment.}
    \label{fig:CC_diagram}
\end{figure}

\section{Experiments}
\label{sec:experiments}

\paragraph{Simulated Environments.} We conduct our experiments in two text-based simulators. 

\textit{CoinCollector}~\cite{yuan2019counting} is a navigation-oriented game where an agent looks for a coin in a set of rooms connected by doors or directly (Figure~\ref{fig:CC_diagram}). Initially, the agent only receives information about the room where it is (e.g., ``You are in the kitchen. To the East you see a closed plain door.''). The global layout, door status, and the coin location are \emph{unknown} to the agent. Thus, the agent should explore different rooms via actions including \texttt{open door} or \texttt{move} to obtain new observations and look for the coin. The task ends successfully when ``coin'' appears in the observation. Otherwise, the task is aborted when a maximum number of actions have been executed. We consider a spectrum of difficulty, ranging from 3 rooms to 11 (the maximum) rooms. We use CoinCollector to study whether a planning agent can construct and maintain a map while exploring a location, as a proof-of-concept to applications like autonomous driving.

\begin{figure}[t!]
    \centering
    \includegraphics[width=\columnwidth]{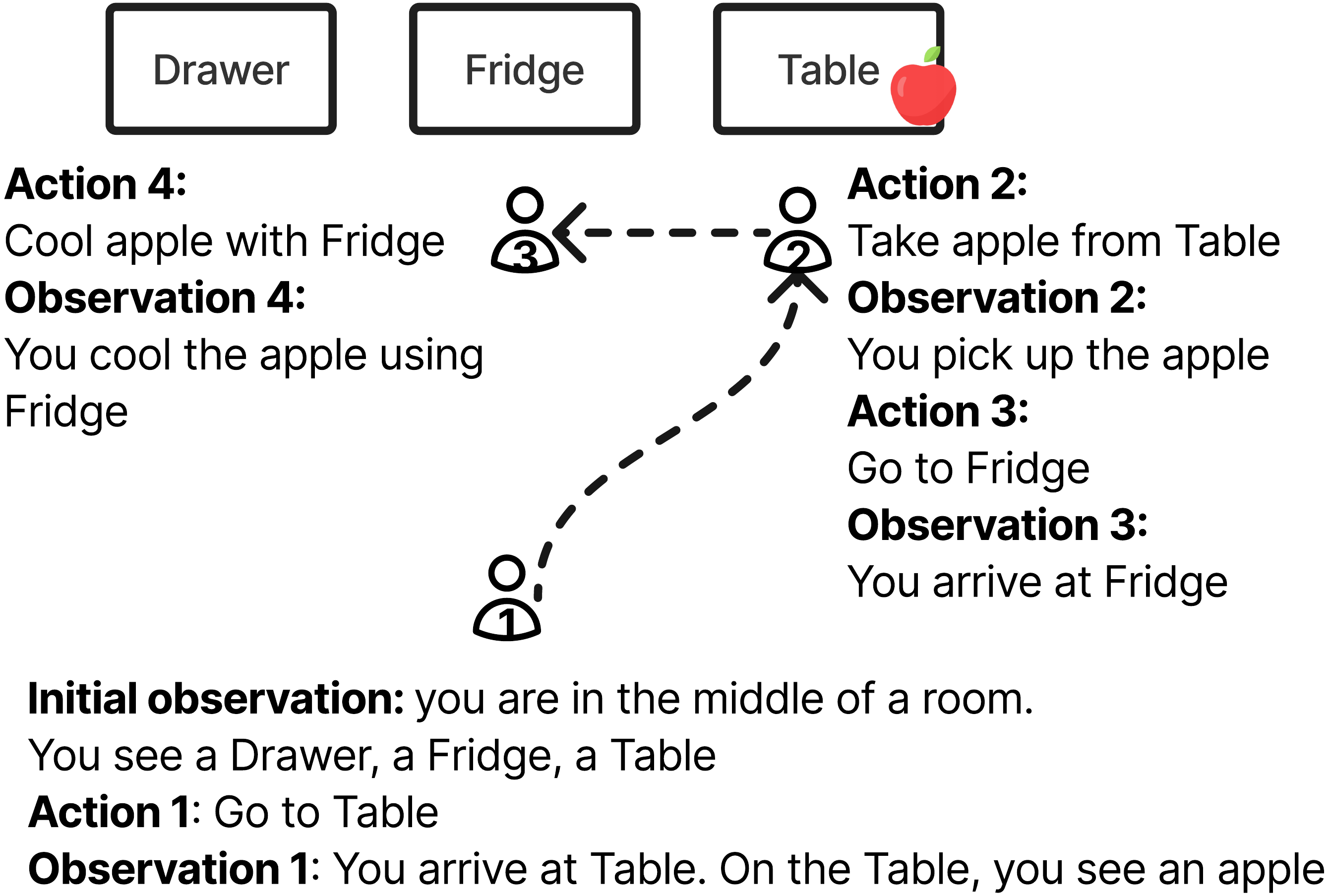}
    \caption{An illustration of ALFWorld.}
    \label{fig:AlfW_diagram}
\end{figure}

\textit{ALFWorld}~\cite{ALFWorld20} is a more complex simulation involving object manipulation in addition to navigation (Figure~\ref{fig:AlfW_diagram}).  Unlike CoinCollector, each instance specifies a composite goal such as ``Put a hot slice of tomato in countertop''.  Besides spatial exploration (moving between receptacles), the agent must manipulate objects via actions including \texttt{pick}, \texttt{put}, \texttt{heat}, \texttt{cool}, \texttt{clean}, and \texttt{slice}.  Partial observability manifests in two ways: (i)~contents in receptacles are hidden (e.g., \texttt{apple} in a \texttt{drawer}) until the receptacle is opened; (ii)~successive state changes (e.g.\,from \texttt{raw} to \texttt{heated}) are visible only via textual feedback after the corresponding action is executed. We use ALFWorld to further study whether a planning agent can propose and refine more complex action semantics which might not always be intuitive, as a proof-of-concept to applications like household robots.

\paragraph{Methods.} We consider the following methods:

\texttt{PlanGen} is \lap, where LLM produces a plan without generating explicit domain and problem definitions (Figure~\ref{fig:baseline}). The method is essentially a combination of ReAct \cite{yao2023reactsynergizingreasoningacting} and self-refining \cite{madaan2023selfrefineiterativerefinementselffeedback}. We experiment with generating a sequence of actions or one action at a time, but only report the latter following \citet{zhang-etal-2024-pddlego}. If the plan cannot be fully executed, the simulation error is returned to LLM to regenerate the plan up to 5 times.

\texttt{PDDLego} \cite{zhang-etal-2024-pddlego} is another baseline using \laf to predict \df and \pf without error refinement. Trial will be aborted if the first generated \pf and \df cannot result in a plan or the action cannot be executed.

\method is our proposed method involving \laf, which generates a PDDL representation (Figure~\ref{fig:pddlegoplus}), thus delegating plan search to a search-based planner (Fast Downward). The representation (\df and \pf) is refined based on solve or simulation error messages up to 5 retries per error, and updated based on new observations. This design provides formal guarantees on plans found by the solver, while still leveraging the LLM flexibility to infer the symbolic representation.

\begin{figure*}[ht]
    \centering
    \includegraphics[width=\textwidth]{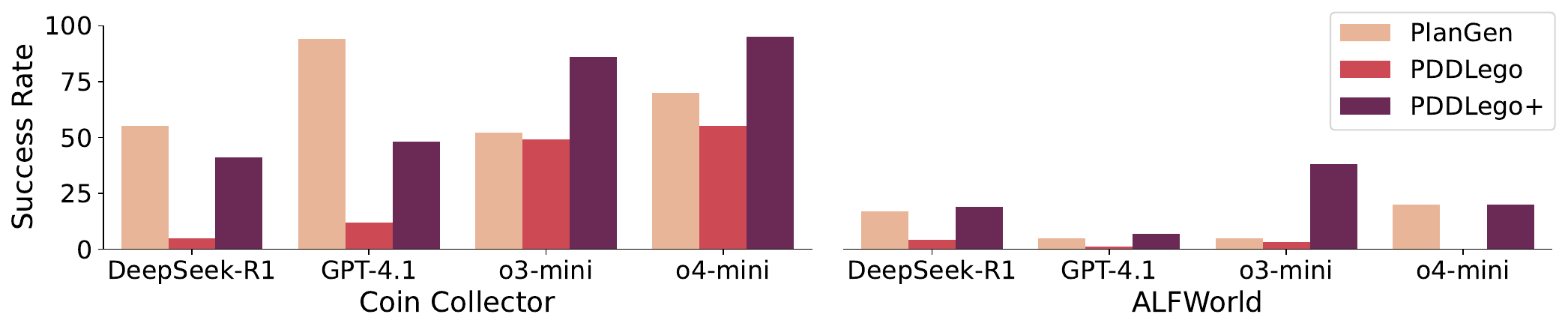}
    \caption{Success rate of two baselines \texttt{PlanGen} and \texttt{PDDLego} and our method \method across four models. \method shows higher success in 6 out 8 model-simulation combinations. In the more challenging ALFWorld, \method outperforms \texttt{PlanGen} for every model.}
    \label{fig:sr_all}
\end{figure*}

\paragraph{Models.} We employ multiple LLMs in our experiments, including both open-source and closed-source models.
\begin{itemize}

\item \texttt{DeepSeek-R1}-671B, a model trained to generate deep chain-of-thought during inference, aimed at multi-step reasoning tasks. 

\item \texttt{GPT-4.1}-2025-04-14, a state-of-the-art, large OpenAI model as of 2025 that does not explicitly generate reasoning tokens by default. 

\item \texttt{o3-mini}-2025-01-31 and \texttt{o4-mini}-2025-04-16, a compact version of OpenAI's flagship reasoning model, also trained to generate a deep chain-of-thought during inference. We use a medium reasoning effort.
\end{itemize} 

We experiment with other models such as \texttt{QwQ-32B}, \texttt{Llama-3.1-70B}, \texttt{GPT-4o-mini},  and \texttt{DeepSeek-R1-Distill-Qwen-32B}, but found they perform significantly worse (in line with the findings of \citet{huang2025limitlanguagemodelsplanning}), so we omit them. (see detailed performance in Appendix C)
\paragraph{Goal Specification.}

In a partially observable environment, the goal cannot be achieved until sufficient exploration. Therefore, the ability of decomposing the goal into an achievable sub-goals is critical for \laf. Unlike \citet{zhang-etal-2024-pddlego} who manually provided the sub-goal in PDDL, we consider two styles of goal specification in the LLM sub-goal generation prompt.

The \textit{simple} prompt merely outlines a coarse recipe (e.g., ``search for the target'' then ``use it to finish the task''), with informal sub-goals. For example, in using the object to finish the task, the sub-goals are ``pick up the target object'', ``move to appropriate receptacle'', and ``interact with the relevant objects and receptacles''. The prompt also encourages exploration of unvisited locations, attempting available actions, and decomposing tasks.

The \textit{detailed} prompt provides a concrete PDDL goal template, e.g., \texttt{(:goal (at ?location))} for CoinCollector or \texttt{(:goal (opened ?receptacle))} for ALFWorld. Even with the prior given by template, the model still needs to decide every sub-goal by filling in the placeholders (\texttt{?location}, \texttt{?obj}) in a way that remains consistent with its current world model. We default to the detailed prompt.

\section{Results and Observations}

We evaluate each LLM over 100 trials with different configurations on each dataset, except for \texttt{o4-mini} evaluated with 20 trials (detailed results in Appendix C). Below, we first report general success rate followed by analysis on complexity, error rate and error fixing performance, prompt ablation, and re-usability of domain knowledge.

\subsection{\method is effective in most cases}

\begin{figure}[t]
    \centering
    \includegraphics[width=\linewidth]{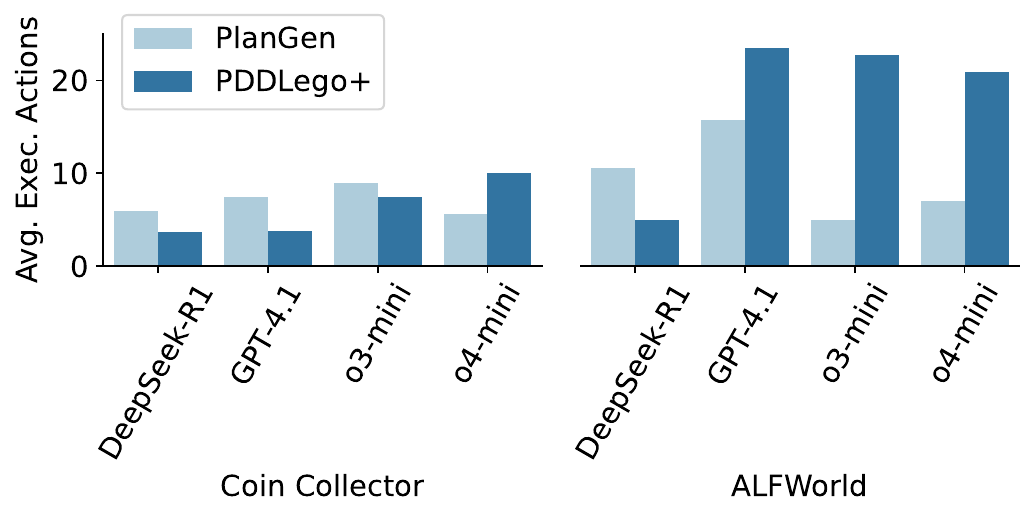}
    \caption{Average number of successful actions executed per trial in CoinCollector and ALFWorld across four models. 
    Successful actions differ from successful steps, because each step can comprise multiple actions. }
    \label{fig:avg_step_all}
\end{figure}

Figures~\ref{fig:sr_all} shows that \method achieves higher success in 6 out of 8 model–environment combinations. Compared to \texttt{PDDLego}, which often fails at the first planning step due to the absence of refinement, \method is consistently more successful across all models and environments.
Compared to \texttt{PlanGen}, \method is not uniformly superior across all backbones; however, this pattern is expected given that \texttt{PlanGen} directly generates actions, while \method requires accurate symbolic formalization before planning.
Importantly, in the more challenging ALFWorld benchmark, \method outperforms \texttt{PlanGen} for \emph{every} evaluated model, and is the only method to solve tasks requiring multi-stage action semantics such as \texttt{clean}, \texttt{heat}, or \texttt{slice}.
We emphasize that our goal is not to claim universal dominance over LLM-as-planner approaches, but to demonstrate the feasibility and advantages of iterative LLM-as-formalizer methods in partially observable environments.

\subsection{\method is robust to complexity}

\begin{figure}[t]
    \centering
    \includegraphics[width=\linewidth]{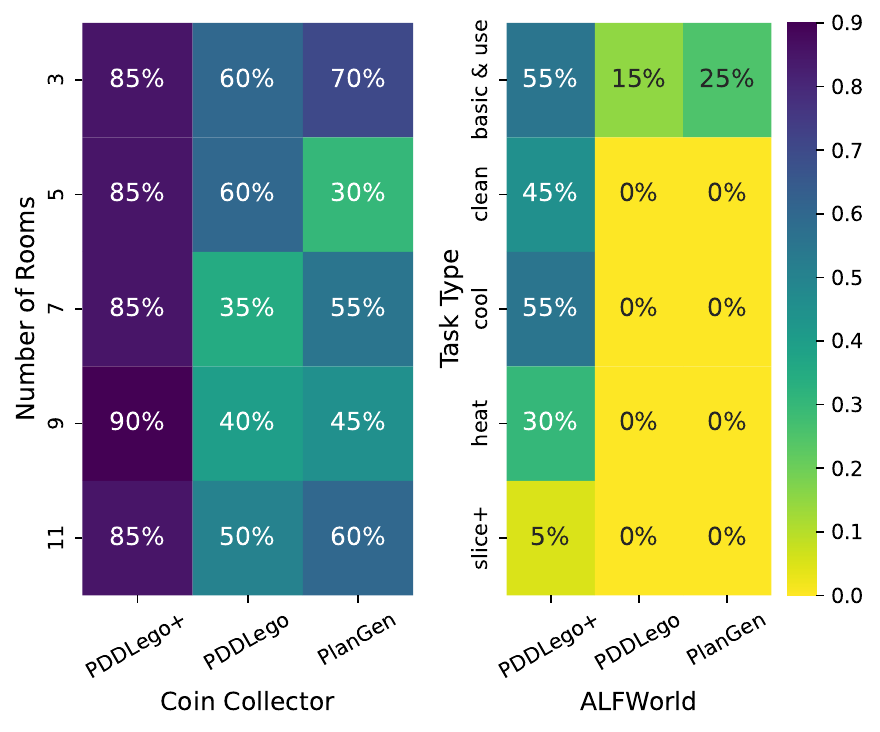}
    \caption{Success rate by difficulty applying \texttt{o3-mini} across \texttt{PlanGen}, \texttt{PDDLego}, and \method.
\textbf{Left:} CoinCollector, binned by number of rooms. \textbf{Right:} ALFWorld, binned by required action \emph{types}}
    \label{fig:breakdown}
\end{figure}

In both simulations, each trial varies by the number of entity space and sometimes action space. To study performance by complexity, in CoinCollector, complexity is dominated by the \emph{entity space}: more rooms imply a larger search graph and, typically, longer paths. We therefore stratify the 100 CC trials into five bins by the number of rooms $\{3,5,7,9,11\}$ (20 trials each). In ALFWorld, every instance contains roughly the same number of locations, but the \emph{action space} varies widely.  We group trials by the \emph{set of action \emph{types}} required to reach the goal, resulting in five ordinal levels: basic actions (\texttt{move}, \texttt{open}, \texttt{use}), plus \texttt{clean}, \texttt{cool}, \texttt{heat}, and everything else like \texttt{slice}. 

While Figure~\ref{fig:sr_all} suggests \method's comparative robustness on the more complex ALFWorld than the simpler CoinCollector, Figure~\ref{fig:breakdown} suggests the same within each simulation. On CoinCollector, as the number of locations from rooms increases, success rate of \method remains constant while that of \texttt{PlanGen} and \texttt{PDDLego} falls gradually. We speculate that once an agent masters the ``open-door + move'' pattern, the added rooms introduce little overhead. On ALFWorld, \texttt{PlanGen} and \texttt{PDDLego} excel only on the easiest category involving basic actions like \textit{use} but fail completely on the rest, whereas \method remains relatively robust, being the only approach to achieve non-zero performance on the hardest category. 

\subsection{Errors are interpretable and fixable}

\begin{figure}[t]
    \centering
    \includegraphics[width=\linewidth]{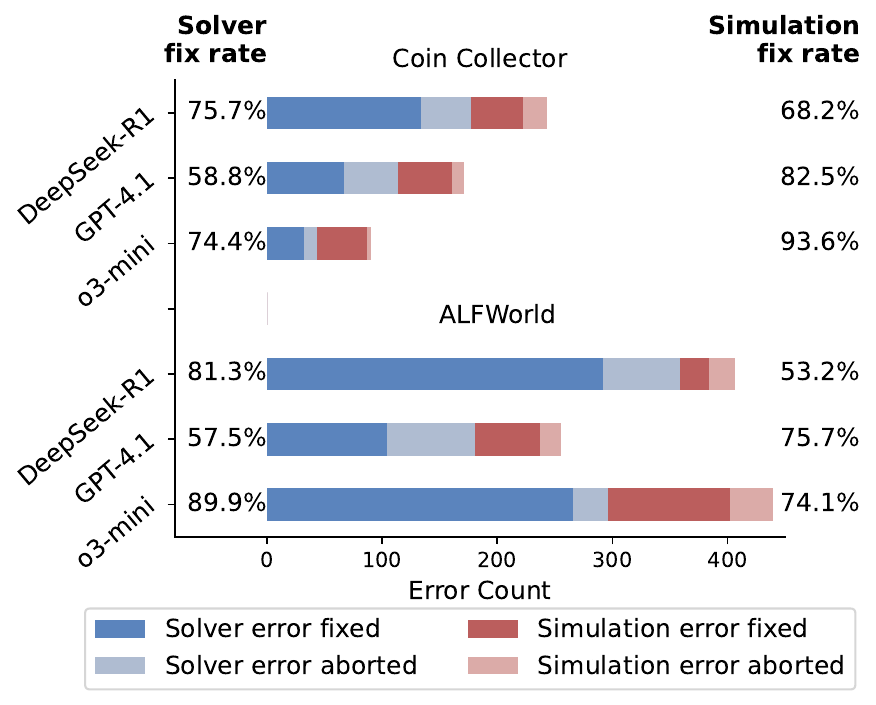}
    \caption{Solver (blue) and simulation (red) error counts with fix and abortion rate of \method for CoinCollector and ALFWorld across three models. \texttt{o3-mini} paired with \method fixes most errors it produces.}
    \label{fig:sim_sol_error}
\end{figure}

Unlike the \texttt{PlanGen}, the \method generates fully determines the resulting plan; such is the nature of \laf methods.
Thus, the reason of any failure can be attributed to the errors in the PDDL in an interpretable manner, allowing causal error analysis that is otherwise impossible for \lap methods. Figures~\ref{fig:sim_sol_error} shows the breakdown every error of \method paired with 3 LLMs in the 100 CoinCollector and 100 ALFWorld trials. At most one solver error or one simulation error is counted per step. Whenever the inner- or outer-loop refines the PDDL so that both the solver the simulator succeed, we mark that error as fixed. If the limit on retries is reached in any loop, the trial terminates and the last unresolved error is labeled as aborted. 
There are far more errors on ALFWorld than on CoinCollector, explained by the former's richer action semantics and longer plans required. We observe that most errors are solver errors for both environments, suggesting the primary obstacle is therefore producing valid PDDL syntactically rather than semantically. Even finer-grained error analysis is performed in Error Analysis Section.

Comparing LLMs, \texttt{o3-mini} with \method fixes most produced errors, and produces the least errors on CoinCollector where all models have reasonable performance. While it encounters most errors on ALFWorld, it is one of the only model-method combination that shows reasonable performance, so the errors are likely due to longer plans. \texttt{DeepSeek-R1} shows a comparable fix rate, though it produces more errors in total; \texttt{GPT-4.1} struggles to recover from solver errors about half of the time.




\subsection{Domain knowledge can be reused}

\begin{figure}[t]
    \centering
    \includegraphics[width=\linewidth]{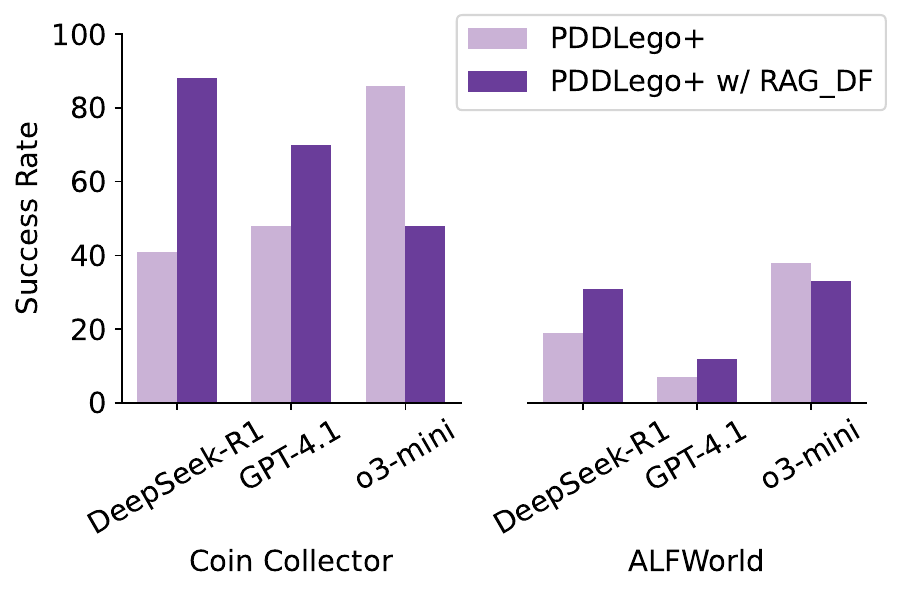}
    \caption{Success rate on CoinCollector and ALFWorld from \method with fixed \df across 3 models.}
    \label{fig:df_rag}
\end{figure}


One beneficial byproduct of \method is the domain knowledge (\df), learned at the end of each trial (see an example in Appendix D). The \df that captures the environment is highly interpretable and can be refined by human practitioners. Even without manual interference, we hypothesize that this \df may be reused to help with future tasks. Therefore, we run a control experiment referred to as retrieval-augmented generation (RAG) with \df. From existing successful \method runs with \texttt{o3-mini}, we randomly retrieve ten \df for environment and hold them fixed for future trials, where the LLMs now only predict \pf. All other settings such as PDDL refinement stay the same.

Figure~\ref{fig:df_rag} supports our hypothesis for \texttt{DeepSeek-R1} and \texttt{GPT-4.1}, but not for \texttt{o3-mini}. Inspecting ten retrieved CoinCollector \df, the 3 complete \df achieve 20/30 successes (67\%), while the 5 missing \texttt{?from} in \texttt{open-door} achieve 20/50 (40\%), and the 2 missing \texttt{door closed} in \texttt{move} achieve 8/20 (40\%). As expected, complete \df perform better, though the gap is smaller than anticipated, warranting further study.


\section{Error Analysis}
\label{sec:error_analysis}

\begin{figure}[t]
    \centering
    \includegraphics[width=0.9\columnwidth]{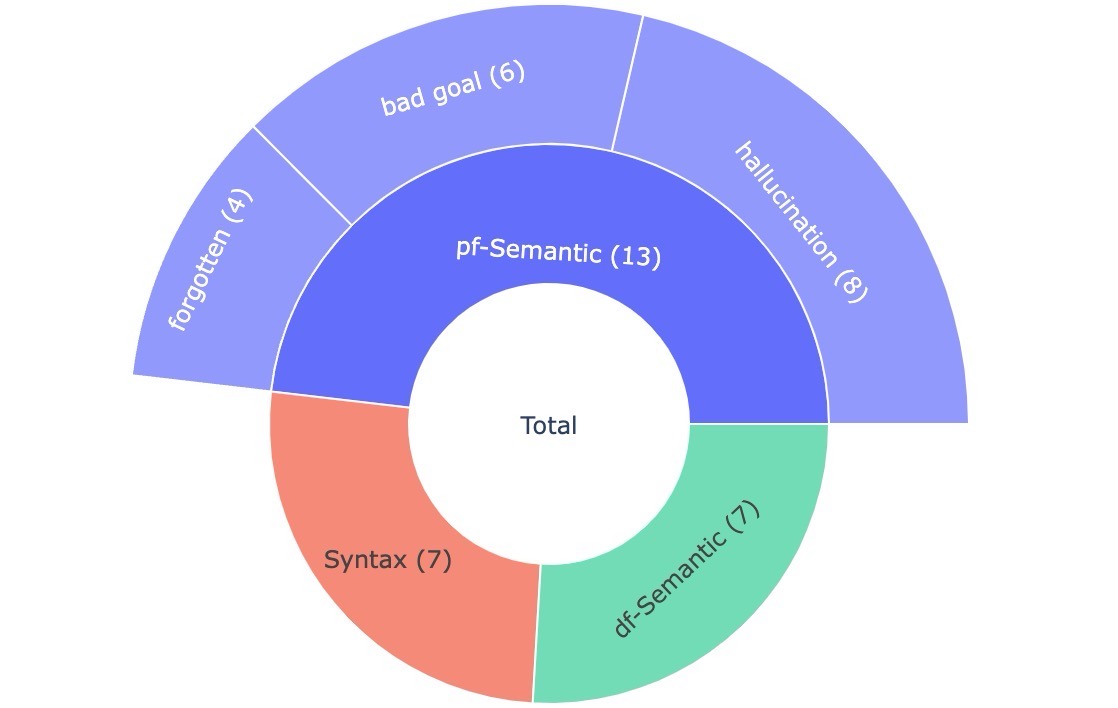}
    \caption{Error breakdown for randomly selected errors in ALFWorld trials with \method and \texttt{o3-mini}, grouped into syntax errors and semantic errors in \df and \pf, further split into fine-grained categories.}
    \label{fig:error_pie}
\end{figure}

Despite \method's improvements, several recurring problems remain. We have three types of errors: syntax error for \df and \pf; semantics error for \df and semantics error for \pf. Based on 27 errors from randomly sampled failed trials from \texttt{o3-mini} + \method, we manually checked what error those trials encounter and remain unresolved after error fixing process. From Figure~\ref{fig:error_pie}, roughly one-third are ordinary syntax or typing slips, and another third from wrong pre- or post-conditions in the \df. The dominant errors are from semantic flaws in \pf. The model either hallucinates unseen facts, keeps or invents an impossible goal, or simply forgets previously observed information. 

\paragraph{Syntax errors in \df and \pf.}
Missing \texttt{:types} lines or mismatched parameter names still happen, yet they are relatively easy: the planner returns a clear error and the inner loop usually fixes them.
\vspace{-\parskip}
\paragraph{Action semantics errors in \df}
Most simulation errors come from missing or wrong preconditions or effects, such as \ omitting \texttt{(holding ?o)} before \texttt{PutObject}. While sometimes fixable by solver error messages, many issues remain dormant and may surface during later execution. Even with successful trials, the action semantics errors in \df also exist if they are never exposed.
\vspace{-\parskip}
\paragraph{Hallucinated facts in \pf.}
Sometimes, non-existent objects or relations are included in the PDDL. For example, the LLM sometimes ``invents'' objects or relations, e.g., it adds \texttt{(in winebottle cabinet1)} when no \texttt{winebottle} was ever observed, despite us prompting “only generate what you see.” In this case, the solver finds a plan that later fails in the simulator. 
\vspace{-\parskip}
\paragraph{Bad goals in \pf.}
Most solver crashes are due to empty, replicated, or impossible goals, such as \texttt{(exists (?to - receptacle) (not (at toilet1)))} or a find object goal even after the object is held. Our detailed prompt reduces but does not eliminate these wrong goals, and the model may keep regenerating the same unreachable target.
\paragraph{Inconsistent world state in \pf.}
Sometimes, previously observed facts are dropped in later stage. For example, it “forgets’’ a room, because LLM overwrites rather than appends to the problem file, even despite we prompting the models not to do so. Stemming from this issue, some trials still terminate after repeated rediscovery of the same area.

\section{Conclusion}

We propose \method, a framework that enables LLMs without extra training to plan in partially observable environments. We use LLMs as a PDDL formalizer that can update its representation based on new observations, and fix it based on error feedback. 
While \method does not dominate \texttt{PlanGen} approaches in all settings, our results establish that iterative symbolic formalization is viable, interpretable, and competitive in partially observable environments, providing a complementary paradigm.

\section*{Limitations}

Despite our framework’s promising results, it comes with simplifying assumptions based on the two simulated environments we experiment with. First, \method requires informative error messages from the simulation to drive the refinement loops, which might not be readily available in all environments. Second, although \method relies on only minimal prior knowledge, its dependence on environment-specific prompt wording could limit its ability to generalize to unseen domains. Third, the success of \method requires powerful, high-capacity LLMs (e.g., \texttt{DeepSeek-R1}, \texttt{o3-mini}) and makes multiple model calls at each step, which is both time- and compute-intensive.

To address those limitations, we identify three key directions to strengthen and generalize our framework. First, we recommend broadening the evaluation by applying the method to additional text-based and vision-based simulated environments with some approach to automatically synthesize error feedback if it is not readily available. Second, we recommend working towards unified, structured memory that is agonistic to domains and problems that reduces the need of any environment-specific guidance. Third, we call for research on improving smaller LLMs' ability to generate formal planning languages like PDDL and increase the efficiency when doing so.

\section*{Acknowledgement}

This research is funded through the National Science Foundation’s Civil Infrastructure Systems (CIS) program, award number 2409847. Any errors or omissions remain the sole responsibility of the authors.

\bibliography{anthology, custom}

\appendix

\lstdefinestyle{appendixpy}{
  basicstyle=\ttfamily\footnotesize,
  numbers=left,
  numberstyle=\tiny,
  stepnumber=1,
  numbersep=6pt,
  frame=single,
  breaklines=true,
  columns=fullflexible,
  keepspaces=true,
  showstringspaces=false
}

\section*{Appendix A: Pseudocode}
\begin{lstlisting}[language=Python,style=appendixpy, numbers=none]
def method_framework(env, LLM, solver, max_time_steps, max_outer_iters, max_inner_iters):
    # Initialize state from environment and create first observation
    state = env.reset()
    obs = env.observe(state)
    goal = env.get_goal()  # goal in natural language

    # Initialize time step
    i = 0

    # Continue until goal is reached or max steps exhausted
    while not env.is_goal(state, goal) and i < max_time_steps:

        # Initial PDDL generation for this time step
        df, pf = LLM.generate(obs)

        # === OUTER LOOP: Based on simulation errors ===
        outer_attempt = 0
        while outer_attempt < max_outer_iters:

            inner_attempt = 0

            # === INNER LOOP: Based on solver errors ===
            while inner_attempt < max_inner_iters:
                plan, err_sol = solver.solve(df, pf)

                if plan is not None:
                    break  # solver succeeded
                else:
                    # Refine using solver error
                    df, pf = LLM.refine_solver(err_sol, df, pf)
                    inner_attempt += 1

            # If no valid plan from solver, exit and fail
            if plan is None:
                print("Solver failed to produce a valid plan.")
                return False

            # Try executing the plan
            success, exec_err = env.execute_plan(plan, state)

            if success:
                # Advance to next state and observation
                state = env.current_state()
                obs = env.observe(state)
                i += 1  # Next time step
                break  # Exit outer loop
            else:
                # Simulation error occurred -> refine entire PDDL
                df, pf = LLM.refine_simulation(exec_err, df, pf)
                outer_attempt += 1

        # If simulation still fails after many retries
        if outer_attempt >= max_outer_iters:
            print("Simulation refinement failed; aborting.")
            return False

    # Return success if goal was reached
    return env.is_goal(state, goal)
\end{lstlisting}

\section*{Appendix B: Reproducibility Details}
\subsection*{B.1: Hardware \& Software Environment}
All experiments were conducted using:
\begin{itemize}
  \item \textbf{OS \& Hardware}: 8*H100 (with in total around 80 hours for all the experiments)
  \item \textbf{Python}: v3.10
  \item \textbf{Core libraries}: requests, textworld\_express, fast-downward-textworld==20.6.2rc1, alfworld, openai, pandas.
  \item \textbf{Setup Instructions}: Included in README.md; recommend using:
  \begin{verbatim}
conda create -n pddlego-plus python=3.10
pip install -r requirements.txt
export OPENAI_API_KEY, deepseek_API
  \end{verbatim}
\end{itemize}

\subsection*{B.2: Random Seed \& Determinism Protocol}
Reproducibility is ensured via explicit seeding:
\begin{itemize}
  \item \textbf{CoinCollector}: For each setting of `num\_locations`, trials use seeds \{0,1,2,3,4\}.
  \item \textbf{ALFWorld}: Game instances were manually selected and considered deterministic (no additional seed variation).
  \item \textbf{LLM prompts \& solvers}: A fixed seed 42 is passed to randomness where applicable (e.g., internal RNGs, API logit bias, solver options).
\end{itemize}

\subsection*{B.3: Preprocessing \& Code Appendix}
All implementation and analysis scripts are included in the repository (see README.md):
\begin{itemize}
  \item \textbf{Core scripts and Preprocessing}: interactive\_CoinCollector.py, interactive\_Alfworld.py
  \item \textbf{Analysis}: results\_parse\_final.py, extract\_actions.py
  \item \textbf{Visualization}: visualize\_pddlego+.ipynb
\end{itemize}

\subsection*{B.4: Number of Runs \& Statistical Reporting}
\begin{itemize}
  \item Each result aggregates over \textbf{100 independent trials} per model/environment (20 for o4-mini).
  \item While we do not perform formal statistical tests, all variability and sample size details are clearly documented.
\end{itemize}

\section*{Appendix C: Trial Results}
\label{sec:trial_results}

The comprehensive results of all experiments are shown in the following tables.

\begin{table*}[ht]
\centering
\small
\resizebox{\linewidth}{!}{
\setlength{\tabcolsep}{3pt}
\begin{tabular}{lccccccccc}
\toprule
 & \multicolumn{3}{c}{DeepSeek-R1} & \multicolumn{3}{c}{GPT-4.1} & \multicolumn{3}{c}{o3-mini} \\
\cmidrule(lr){2-4} \cmidrule(lr){5-7} \cmidrule(lr){8-10}
Metric                     & PlanGen & PDDLego & PDDLego+ & PlanGen & PDDLego & PDDLego+ & PlanGen & PDDLego & PDDLego+  \\
\midrule
trial\_count               & 100     & 100     & 100      & 100     & 100     & 100      & 100     & 100     & 100    \\
succeed\_count             & 55      & 5       & 41       & 94      & 12      & 48       & 52      & 49      & 86     \\
success\_rate              & 55\%    & 5\%     & 41\%     & 94\%    & 12\%    & 48\%     & 52\%    & 49\%    & 86\%   \\
total\_solver\_errors      & –       & –       & 177      & –       & –       & 114      & –       & –       & 43     \\
total\_solver\_fixed       & –       & –       & 134      & –       & –       & 67       & –       & –       & 32       \\
solver\_error\_fix\_rate   & –       & –       & 76\%     & –       & –       & 59\%     & –       & –       & 74\%    \\
total\_simulation\_errors  & 161     & –       & 66       & 174     & –       & 57       & 185     & –       & 47       \\
total\_simulation\_fixed   & 112     & –       & 45       & 153     & –       & 47       & 138     & –       & 44      \\
simulation\_error\_fix\_rate & 70\%  & –       & 68\%     & 88\%    & –       & 82\%     & 75\%    & –       & 94\%   \\
total\_abort\_solver       & –       & –       & 43       & –       & –       & 47       & –       & –       & 11      \\
total\_abort\_simulation   & 49      & –       & 21       & 21      & –       & 10       & 47      & –       & 3        \\
avg\_steps\_success        & 6.1     & 1.3     & 3.1      & 7.4     & 2.4     & 3.0      & 4.4     & 2.8     & 3.4     \\
avg\_steps\_success*       & –       & –       & 3.63     & –       & –       & 3.77     & –       & –       & 7.48   \\
avg\_steps\_failure        & 5.9     & 2.5     & 2.1      & 20.0    & 2.5     & 2.0      & 8.9     & 6.9     & 6.4    \\
\midrule
 & \multicolumn{3}{c}{o4-mini} & \multicolumn{3}{c}{Qwen}  \\
 \cmidrule(lr){2-4} \cmidrule(lr){5-7}
Metric                     & PlanGen & PDDLego & PDDLego+ & PlanGen & PDDLego & PDDLego+ \\
\cmidrule(){1-7}
trial\_count  & 20      & 20      & 20       & 100     & 100     & 100      \\
succeed\_count  & 14      & 11      & 19       & 46      & 0       & 0      \\
success\_rate    & 70\%    & 55\%    & 95\%     & 46\%    & 0\%     & 0\%      \\
total\_solver\_errors      & –       & –       & 8        & –       & –       & 100      \\
total\_solver\_fixed & –       & –       & 8        & –       & –       & 0  \\
solver\_error\_fix\_rate & –       & –       & 100\%    & –       & –       & 0\%      \\
total\_simulation\_errors & 27      & –       & 5        & 182     & –       & 0   \\
total\_simulation\_fixed & 18       & –       & 4        & 117       & –       & 100      \\
simulation\_error\_fix\_rate   & 67\%    & –       & 80\%     & 64\%    & –       & –        \\
total\_abort\_solver   & -      & –       & 0        & -     & –       & 0       \\
total\_abort\_simulation   & 9       & –       & 1        & 65      & –       & 0        \\
avg\_steps\_success    & 5.6     & 4.2     & 4.1      & 5.2     & –       & –        \\
avg\_steps\_success*      & –       & –       & 10       & –       & –       & –    \\
avg\_steps\_failure        & 5.8     & 5.9     & 6.0      & 6.5     & –       & –        \\
\bottomrule
\end{tabular}
}
\caption{CoinCollector – 100-trial averages. Three planning frameworks per LLM; dashes mark metrics that are not defined for that framework.}
\label{tab:cc_full_metrics}
\end{table*}

\begin{table*}[ht]
\centering
\small
\setlength{\tabcolsep}{3pt}
\resizebox{\linewidth}{!}{
\begin{tabular}{l ccc ccc ccc}
\toprule
 & \multicolumn{3}{c}{DeepSeek-R1} & \multicolumn{3}{c}{GPT-4.1} & \multicolumn{3}{c}{o3-mini} \\
\cmidrule(lr){2-4} \cmidrule(lr){5-7} \cmidrule(lr){8-10}
Metric                     & PlanGen & PDDLego & PDDLego+ & PlanGen & PDDLego & PDDLego+ & PlanGen & PDDLego & PDDLego+ \\
\midrule
trial\_count               & 100     & 100     & 100      & 100     & 100     & 100      & 100     & 100     & 100      \\
succeed\_count             & 17      & 4       & 19       & 5       & 1       & 7        & 5       & 3       & 38       \\
success\_rate              & 17\%    & 4\%     & 19\%     & 5\%     & 1\%     & 7\%      & 5\%     & 3\%     & 38\%     \\
total\_solver\_errors      & –       & –       & 359      & –       & –       & 181      & –       & –       & 296      \\
total\_solver\_fixed       & –       & –       & 292      & –       & –       & 104      & –       & –       & 266      \\
solver\_error\_fix\_rate   & –       & –       & 81\%     & –       & –       & 57\%     & –       & –       & 89\%     \\
total\_simulation\_errors  & 97      & –       & 47       & 112     & –       & 74       & 212     & –       & 143      \\
total\_simulation\_fixed   & 83      & –       & 25       & 93      & –       & 56       & 154     & –       & 106      \\
simulation\_error\_fix\_rate & 85\%  & –       & 53\%     & 83\%    & –       & 75\%     & 72\%    & –       & 74\%     \\
total\_abort\_solver       & –       & –       & 67       & –       & –       & 77       & –       & –       & 30       \\
total\_abort\_simulation   & 14      & –       & 22       & 19      & –       & 18       & 58      & –       & 37       \\
avg\_steps\_success        & 10.6    & 1.3     & 9.0      & 15.7    & 17.0    & 30.6     & 5.0     & 2.0     & 19.2     \\
avg\_steps\_success*       & –       & –       & 4.92     & –       & –       & 23.5     & –       & –       & 22.75    \\
avg\_steps\_failure        & 18.9    & 5.1     & 4.0      & 18.4    & 24.3    & 23.7     & 11.6    & 21.5    & 21.2     \\
\midrule
 & \multicolumn{3}{c}{o4-mini} & \multicolumn{3}{c}{Qwen} \\
\cmidrule(lr){2-4} \cmidrule(lr){5-7}
Metric                     & PlanGen & PDDLego & PDDLego+ & PlanGen & PDDLego & PDDLego+ \\
\cmidrule(){1-7}
trial\_count               & 20      & 20      & 20       & 88       & 100     & 100      \\
succeed\_count             & 4       & 0       & 4        & 3       & 0       & 0        \\
success\_rate              & 20\%    & 0\%     & 20\%     & 3\%       & 0\%     & 0\%      \\
total\_solver\_errors      & –       & –       & 62       & –       & –       & 100      \\
total\_solver\_fixed       & –       & –       & 51       & –       & –       & 0        \\
solver\_error\_fix\_rate   & –       & –       & 82\%     & –       & –       & 0\%      \\
total\_simulation\_errors  & 20      & –       & 16       & 180       & –       & 0        \\
total\_simulation\_fixed   & 16      & –       & 13       & 102       & –       & 100      \\
simulation\_error\_fix\_rate & 80\%  & –       & 81\%     & 56\%       & –       & –        \\
total\_abort\_solver       & –       & –       & 11       & –       & –       & 0        \\
total\_abort\_simulation   & 4       & –       & 3        & 78       & –       & 0        \\
avg\_steps\_success        & 7.0     & –       & 15.0     & 13.3       & –       & –        \\
avg\_steps\_success*       & –       & –       & 20.95    & –       & –       & –        \\
avg\_steps\_failure        & 19.6    & 21.0    & 21.5     & 4.7       & –       & –        \\
\bottomrule
\end{tabular}
}
\caption{ALFWorld – 100-trial averages. Three planning frameworks per LLM; dashes mark metrics that are not defined for that framework.}
\label{tab:alfw_full_metrics}
\end{table*}

\begin{table}[t!]
\centering
\small
\setlength{\tabcolsep}{3pt}
\resizebox{\columnwidth}{!}{
\begin{tabular}{lcccccc}
\toprule
 & \multicolumn{2}{c}{DeepSeek-R1} & \multicolumn{2}{c}{GPT-4.1} & \multicolumn{2}{c}{o3-mini} \\
\cmidrule(lr){2-3}\cmidrule(lr){4-5}\cmidrule(lr){6-7}
Metric                        & simple & detailed & simple & detailed & simple & detailed \\
\midrule
trial\_count                  & 100    & 100      & 100    & 100      & 100    & 100      \\
succeed\_count                & 21     & 41       & 36     & 48       & 68     & 86       \\
success\_rate                 & 21\%   & 41\%     & 36\%   & 48\%     & 68\%   & 86\%     \\
total\_solver\_errors         & 125    & 177       & 115    & 114      & 88     & 43       \\
total\_solver\_fixed          & 54     & 134       & 53     & 67      & 65     & 32        \\
solver\_error\_fix\_rate      & 43\%   & 76\%     & 46\%   & 59\%     & 74\%   & 74\%      \\
total\_simulation\_errors     & 25     & 66       & 35     & 57       & 44     & 47       \\
total\_simulation\_fixed      & 15     & 45       & 27     & 47       & 33     & 44       \\
simulation\_error\_fix\_rate  & 60\%   & 68\%     & 77\%   & 82\%     & 75\%   & 94\%     \\
total\_abort\_solver          & 71     & 43       & 62     & 47       & 23     & 11       \\
total\_abort\_simulation      & 10     & 21        & 8     & 10        & 11     & 3        \\
avg\_steps\_success           & 2.6    & 3.1      & 3.1    & 3.0      & 3.2    & 3.4      \\
avg\_steps\_failure           & 1.2    & 2.1      & 2.1    & 2.0      & 4.7    & 6.4      \\
\bottomrule
\end{tabular}
}
\caption{Success Rate (\%) on 100 trials in CoinCollector: simple prompt vs.\ detailed prompt under \method{}}
\label{tab:cc_goal_compare}
\end{table}

\begin{table}[t!]
\centering
\small
\setlength{\tabcolsep}{3pt}
\resizebox{\columnwidth}{!}{
\begin{tabular}{l cc cc}
\toprule
 & \multicolumn{2}{c}{DeepSeek-R1} & \multicolumn{2}{c}{GPT-4.1} \\
\cmidrule(lr){2-3}\cmidrule(lr){4-5}
Metric                        & simple & detailed & simple & detailed \\
\midrule
trial\_count                  & 100    & 100      & 100    & 100      \\
succeed\_count                & 5      & 19       & 0      & 7       \\
success\_rate                 & 5\%    & 19\%     & 0\%    & 7\%     \\
total\_solver\_errors         & 198    & 359      & 123    & 181      \\
total\_solver\_fixed          & 121    & 292       & 24     & 104      \\
solver\_error\_fix\_rate      & 61\%   & 81\%     & 19\%   & 57\%     \\
total\_simulation\_errors     & 49     & 47       & 15     & 74       \\
total\_simulation\_fixed      & 8      & 25       & 2      & 56       \\
simulation\_error\_fix\_rate  & 16\%   & 53\%     & 13\%   & 75\%     \\
total\_abort\_solver          & 77     & 67       & 99     & 77       \\
total\_abort\_simulation      & 41     & 22       & 13     & 18       \\
avg\_steps\_success           & 1.6    & 9.0     & –      & 30.6     \\
avg\_steps\_failure           & 1.1    & 4.0     & 1.2    & 23.7     \\
\midrule
 & \multicolumn{4}{c}{o3-mini} \\
\cmidrule(lr){2-5}
Metric                        & simple & detailed & simple+goal & simple+hint \\
\midrule
trial\_count                  & 100    & 100      & 100         & 100          \\
succeed\_count                & 7      & 38       & 2           & 2            \\
success\_rate                 & 7\%    & 38\%     & 2\%         & 2\%          \\
total\_solver\_errors         & 449    & 296      & 391         & 405          \\
total\_solver\_fixed          & 435    & 266      & 369         & 387          \\
solver\_error\_fix\_rate      & 96\%   & 89\%     & 94\%        & 95\%         \\
total\_simulation\_errors     & 114    & 143      & 123         & 92           \\
total\_simulation\_fixed      & 27     & 106      & 35          & 7            \\
simulation\_error\_fix\_rate  & 23\%   & 74\%     & 28\%        & 7\%          \\
total\_abort\_solver          & 14     & 30       & 22          & 18           \\
total\_abort\_simulation      & 87     & 37       & 88          & 85           \\
avg\_steps\_success           & 3.4    & 19.2     & 1.0         & 1.0          \\
avg\_steps\_failure           & 3.7    & 21.2     & 2.2         & 1.4          \\
\bottomrule
\end{tabular}
}
\caption{Success Rate (\%) on 100 trials in ALFWorld: simple prompt vs.\ detailed prompt under \method{}}
\label{tab:alf_goal_compare}
\end{table}

\begin{table*}[t!]
\centering
\small
\setlength{\tabcolsep}{3pt}
\begin{tabular}{l ccccc ccccc}
\toprule
 & \multicolumn{5}{c}{DeepSeek-R1} & \multicolumn{5}{c}{GPT-4.1} \\
\cmidrule(lr){2-6}\cmidrule(lr){7-11}
Metric & 3 & 5 & 7 & 9 & 11 & 3 & 5 & 7 & 9 & 11 \\
\midrule
trial\_count          & 20 & 20 & 20 & 20 & 20 & 20 & 20 & 20 & 20 & 20 \\
succeed\_count        & 11 & 9  & 9  & 6  & 6  & 14 & 9  & 12 & 7  & 6  \\
success\_rate         & 55\% & 45\% & 45\% & 30\% & 30\% & 70\% & 45\% & 60\% & 35\% & 30\% \\
total\_solver\_errors & 29 & 41 & 31 & 38 & 38 & 25 & 20 & 22 & 25 & 22 \\
total\_solver\_fixed  & 23 & 34 & 23 & 27 & 27 & 20 & 10 & 15 & 14 &  8 \\
solver\_error\_fix\_rate & 79\% & 83\% & 74\% & 71\% & 71\% & 80\% & 50\% & 68\% & 56\% & 36\% \\
total\_simulation\_errors & 13 & 15 & 12 & 14 & 12 & 10 & 11 & 13 & 15 &  8 \\
total\_simulation\_fixed  &  9 & 11 &  8 &  9 &  8 &  9 &  9 & 12 &  10 &  7 \\
simulation\_error\_fix\_rate & 69\% & 73\% & 67\% & 64\% & 67\% & 90\% & 82\% & 92\% & 67\% & 88\% \\
total\_abort\_solver       &  6 &  7 &  8 & 11 & 11 &  5 & 10 &  7 & 11 & 14 \\
total\_abort\_simulation   &  4 &  4 &  4 &  5 &  4 &  1 &  2 &  1 &  5 &  1 \\
avg\_steps\_success        & 1.7 & 2.6 & 3.0 & 4.3 & 4.0 & 1.6 & 2.8 & 4.3 & 3.0 & 3.0 \\
avg\_steps\_failure        & 1.4 & 1.9 & 2.0 & 2.3 & 2.9 & 1.8 & 1.5 & 1.6 & 3.2 & 1.8 \\
\midrule
 & \multicolumn{5}{c}{o3-mini} & \multicolumn{5}{c}{o4-mini} \\
\cmidrule(lr){2-6}\cmidrule(lr){7-11}
Metric & 3 & 5 & 7 & 9 & 11 & 3 & 5 & 7 & 9 & 11 \\
\cmidrule(){1-11}
trial\_count          & 20 & 20 & 20 & 20 & 20 & 4 & 4 & 4 & 4 & 4 \\
succeed\_count        & 17 & 17 & 17 & 18 & 17 & 4 & 4 & 3 & 4 & 4 \\
success\_rate         & 85\% & 85\% & 85\% & 90\% & 85\% & 100\% & 100\% & 75\% & 100\% & 100\% \\
total\_solver\_errors &  8 &  9 &  7 &  8 & 11 & 1 & 1 & 1 & 0 & 5 \\
total\_solver\_fixed  &  5 &  6 &  4 &  7 & 10 & 1 & 1 & 1 & 0 & 5 \\
solver\_error\_fix\_rate & 63\% & 67\% & 57\% & 88\% & 91\% & 100\% & 100\% & 100\% & - & 100\% \\
total\_simulation\_errors & 1 & 0 & 14 & 15 & 17 & 0 & 1 & 1 & 1 & 2 \\
total\_simulation\_fixed  & 0 & 0 & 12 & 15 & 17 & 0 & 1 & 0 & 1 & 2 \\
simulation\_error\_fix\_rate & 0\% & - & 86\% & 100\% & 100\% & - & 100\% & 0\% & 100\% & 100\% \\
total\_abort\_solver       & 3 & 3 & 3 & 1 & 1 & 0 & 0 & 0 & 0 & 0 \\
total\_abort\_simulation   & 1 & 0 & 2 & 0 & 0 & 0 & 0 & 1 & 0 & 0 \\
avg\_steps\_success        & 1.8 & 2.7 & 3.2 & 3.7 & 5.8 & 1.5 & 2.3 & 4.7 & 3.5 & 8.8 \\
avg\_steps\_failure        & 1.3 & 1.0 & 4.0 & 10.5 & 15.3 & -- & -- & 6.0 & -- & -- \\
\bottomrule
\end{tabular}
\caption{Success Rate (\%) on 100 trials in CoinCollector: five difficulty levels}
\label{tab:cc_difficult_compare}
\end{table*}

\begin{table*}[t!]
\centering
\small
\setlength{\tabcolsep}{3pt}
\begin{tabular}{l ccccc ccccc}
\toprule
 & \multicolumn{5}{c}{DeepSeek-R1} & \multicolumn{5}{c}{GPT-4.1} \\
\cmidrule(lr){2-6}\cmidrule(lr){7-11}
Metric & basic\&use & clean & cool & heat & slice+ & basic\&use & clean & cool & heat & slice+ \\
\midrule
trial\_count          & 20 & 20 & 20 & 20 & 20 & 20 & 20 & 20 & 20 & 20 \\
succeed\_count        &  5 &  4 &  4 &  6 &  0 &  3 &  1 &  1 &  2 &  0 \\
success\_rate         & 25\% & 20\% & 20\% & 30\% & 0\% & 15\% & 5\% & 5\% & 10\% & 0\% \\
total\_solver\_errors & 45 & 84 & 100 & 56 & 74 & 37 & 35 & 35 & 34 & 40 \\
total\_solver\_fixed  & 30 & 70 & 90  & 43 & 59 & 20 & 21 & 22 & 18 & 23 \\
solver\_error\_fix\_rate & 66\% & 83\% & 90\% & 76\% & 79\% & 54\% & 60\% & 62\% & 52\% & 57\% \\
total\_simulation\_errors &  7 & 12 & 16 &  7 &  5 & 15 & 18 & 11 & 14 & 16 \\
total\_simulation\_fixed  &  5 &  7 &  8 &  5 &  0 & 11 & 11 & 10 & 10 & 14 \\
simulation\_error\_fix\_rate & 71\% & 58\% & 50\% & 71\% & 0\% & 73\% & 61\% & 91\% & 71\% & 88\% \\
total\_abort\_solver       & 15 & 14 & 10 & 13 & 15 & 17 & 14 & 13 & 16 & 17 \\
total\_abort\_simulation   &  2 &  5 &  8 &  2 &  5 &  4 &  7 &  1 &  4 &  2 \\
avg\_steps\_success        & 3.0 & 20.3 & 4.0 & 8.8 & \(-\) & 17.3 & 41.0 & 33.0 & 31.0 & \(-\) \\
avg\_steps\_failure        & 4.1 &  1.9 & 6.3 & 2.4 & 5.3 & 15.1 & 20.5 & 33.6 & 18.2 & 31.4 \\
\midrule
 & \multicolumn{5}{c}{o3-mini} & \multicolumn{5}{c}{o4-mini} \\
\cmidrule(lr){2-6}\cmidrule(lr){7-11}
Metric & basic\&use & clean & cool & heat & slice+ & basic\&use & clean & cool & heat & slice+ \\
\cmidrule(){1-11}
trial\_count          & 20 & 20 & 20 & 20 & 20 & 4 & 4 & 4 & 4 & 4 \\
succeed\_count        & 11 &  9 & 11 &  6 &  1 & 1 & 1 & 0 & 2 & 0 \\
success\_rate         & 55\% & 45\% & 55\% & 30\% & 5\% & 25\% & 25\% & 0\% & 50\% & 0\% \\
total\_solver\_errors & 35 & 51 & 60 & 80 & 70 & 7 & 13 & 12 & 14 & 16 \\
total\_solver\_fixed  & 27 & 45 & 54 & 74 & 66 & 5 & 12 &  8 & 14 & 12 \\
solver\_error\_fix\_rate & 77\% & 88\% & 90\% & 93\% & 94\% & 71\% & 92\% & 67\% & 100\% & 75\% \\
total\_simulation\_errors & 21 & 23 & 31 & 39 & 29 & 4 & 5 & 3 & 3 & 1 \\
total\_simulation\_fixed  & 19 & 16 & 25 & 30 & 16 & 3 & 5 & 3 & 1 & 1 \\
simulation\_error\_fix\_rate & 90\% & 70\% & 80\% & 77\% & 55\% & 75\% & 100\% & 100\% & 33\% & 100\% \\
total\_abort\_solver       &  8 &  6 &  6 &  6 &  4 & 2 & 1 & 4 & 0 & 4 \\
total\_abort\_simulation   &  2 &  7 &  6 &  9 & 13 & 1 & 0 & 0 & 2 & 0 \\
avg\_steps\_success        & 8.6 & 15.7 & 31.4 & 27.2 & 13.0 & 5.0 & 23.0 & \(-\) & 17.0 & \(-\) \\
avg\_steps\_failure        & 6.2 & 22.3 & 30.2 & 18.0 & 29.1 & 13.3 & 35.0 & 18.8 & 11.5 & 28.8 \\
\bottomrule
\end{tabular}
\caption{Success Rate (\%) on 100 trials in ALFWorld: five difficulty levels}
\label{tab:alfw_difficult_compare}
\end{table*}

\section*{Appendix D: Trajectory Examples}
\label{sec:trajectory_examples}

We illustrate the trajectory of a successful trial by \method paired with \texttt{o3-min} in CoinCollector. Note how the model initially predicts a flawed \df, which is not exposed until much later, but eventually fixed the error. The \pf is updated after each step with new observations. 

\begin{lstlisting}[frame=single,style=appendixpy, numbers=none, numbers=none]
Observation 1
    You are in the kitchen. To the South you see a closed patio door. To the West you see a closed plain door.

DF 1  
    (define (domain explore)
      (:requirements :strips :typing)
      (:types location direction)
      (:predicates
        (at ?loc - location)
        (door-closed ?loc1 - location ?loc2 - location ?dir - direction)
        (door-open ?loc1 - location ?loc2 - location ?dir - direction)
      )
      (:action open-door
        :parameters (?loc1 - location ?loc2 - location ?dir - direction)
        :precondition (and (at ?loc1) (door-closed ?loc1 ?loc2 ?dir))
        :effect (and (not (door-closed ?loc1 ?loc2 ?dir)) (door-open ?loc1 ?loc2 ?dir))
      )
      (:action move
        :parameters (?from - location ?to - location ?dir - direction)
        :precondition (and (at ?from) (door-open ?from ?to ?dir))
        :effect (and (not (at ?from)) (at ?to))
      )
    ) 
    

PF 1
    (define (problem explore-problem)
      (:domain explore)
      (:objects
        kitchen patio plain - location
        south west - direction
      )
      (:init
        (at kitchen)
        (door-closed kitchen patio south)
        (door-closed kitchen plain west)
      )
      (:goal (at patio))
    ) 
    

SOLVER PLAN
  1. open door to south
  2. move south

EXEC RESULT
  OK, patio door opened, revealing backyard, agent in backyard (= patio)

\end{lstlisting}

\begin{lstlisting}[frame=single,style=appendixpy, numbers=none, numbers=none]
Observation 2
    You open the patio door, revealing the backyard. You are in the backyard. Through an open patio door, to the North you see the kitchen. To the South you see the driveway. To the East you see the street. To the West you see a closed patio door. 

DF 2   (unchanged)

PF 2   *first try causing simulation error*
  :init  (at backyard)
         (door-open kitchen backyard south)
         (*@\textbf{(door-closed backyard driveway south)}@*)
         (door-closed backyard unknown-west west)
  :goal  (at driveway)

SOLVER PLAN
    1. open door to south
    2. move south
    
EXEC RESULT
    ERROR: "No door to open."  (there is no patio door connecting backyard and driveway!)


DF 2.2  *refined by realizeing that not always a door connecting two locations*
    (define (domain explore)
      (:requirements :strips :typing :negative-preconditions :disjunctive-preconditions)
      (:types location direction)
      (:predicates
        (at ?loc - location)
        (door-closed ?loc1 - location ?loc2 - location ?dir - direction)
        (door-open ?loc1 - location ?loc2 - location ?dir - direction)
        (*@\textbf{(door-exists ?loc1 - location ?loc2 - location ?dir - direction)}@*)
        (*@\textbf{(no-door ?loc1 - location ?loc2 - location ?dir - direction)}@*)
      )
      (:action open-door ...)
      (:action move ...)
    ) 

PF 2.2  *pairing with updated domain file*
    :init (at backyard)
          (*@\textbf{(no-door backyard driveway south)}@*)
          (*@\textbf{(no-door backyard street east)}@*)
          ...

SOLVER PLAN
  1. move south

EXEC RESULT
  OK, moved to driveway. (Error fixed!)
\end{lstlisting}

\begin{lstlisting}[frame=single,style=appendixpy, numbers=none]
Observation 3
    You are in the driveway. To the North you see the backyard.

DF 3 (unchanged)

PF 3 
    (:init
        (at driveway)
        (no-door backyard driveway south)
        (no-door backyard street east)
        (*@\textbf{(no-door driveway backyard north)}@*)
      )
      (:goal
        (at street)
      )

SOLVER PLAN
  1. move north
  2. move east

EXEC RESULT
  OK, moved to street.

\end{lstlisting}

\begin{lstlisting}[frame=single,style=appendixpy, numbers=none]
Observation 4
    You are in the street. To the North you see a closed sliding door. To the West you see the backyard. 

DF 4 (unchanged)
    

PF 4
    (:init
        (at street)
        ...
        ;; Sliding door between street and unknown-north (closed)
        (door-exists street unknown-north north)
        (door-exists unknown-north street south)
        (door-closed street unknown-north north)
        (door-closed unknown-north street south)
      )
      (:goal
        (at unknown-north)
      )
    ) 

SOLVER PLAN
  1. open door to north
  2. move north

EXEC RESULT
  OK, moved to the supermarket and saw a coin there! Task accomplished.


SUMMARY
Errors encountered and fixed
   Missing (no-door ...) relation  -> added after sim-error
   All subsequent plans executed without error
Final state
   DF: full action model with door-exists / door-closed / no-door
   PF: map of six rooms, all connectivity relations consistent
   Goal achieved (coin observed) in 4 steps
\end{lstlisting}

\section*{Appendix E: Prompt Design}
\label{sec:prompts}

\begin{figure}[ht]
    \centering
    \includegraphics[width=\columnwidth]{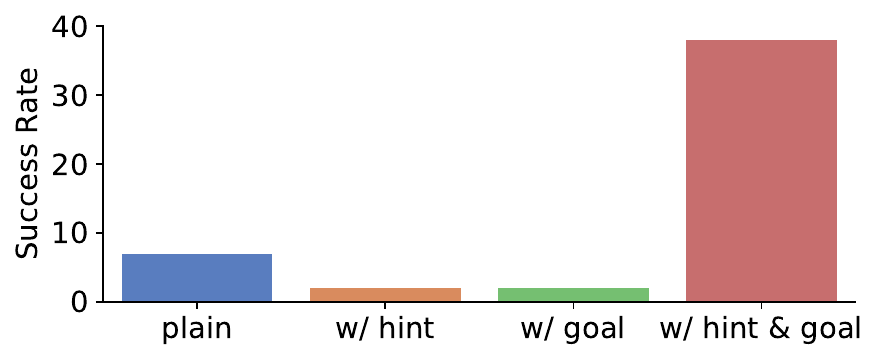}
    \caption{Ablation study of the o3-mini + \method framework on ALFWorld, comparing four prompt variants: plain, plain + hint, plain + goal, and detailed.}
    \label{fig:prompt_sensitivity}
\end{figure}

For the \texttt{PlanGen}, the prompt starts with a single instruction that positions the LLM as a decision maker to generate a plan. A brief header reminds the model that all actions must be strictly grounded in the current textual observation and that it may not invent unseen objects or relations. Next, we provide (i) the natural language goal, (ii) the latest observation, and (iii) the complete list of valid action templates with their parameters (e.g., \texttt{go to [towelholder1]}, \texttt{heat [bread1] with [microwave1]}). In case of a simulation error or a new observation following a successful execution of the plan, the model is prompted again with the trajectory so far and the error (if any) to generate a new plan.

For \texttt{PDDLego} and \method, every trial begins with a single master prompt that tells the LLM it is an expert in PDDL whose only job is to \textit{generate and repair} PDDL from scratch. This is followed by the same reminder, goal, observation, and valid action templates as described above. In case of an error (\method only) or a new observation following a successful execution of the plan (both), the model is prompted again with the trajectory so far, the previous PDDL, and the error (if any) to generate a new pair of \df and \pf.

We obtain \textbf{solver errors} from an online fast downward planner\footnote{\url{https://api.planning.domains/}} which are precise and generally helpful, pinpointing syntactical issues such as missing predicates or unknown types (e.g. \emph{parameter ?SHARP\_O of action SliceObject has unknown or empty type SharpObject}). However, the default \textbf{simulation errors} in both CoinCollector and ALFWorld are often vague and unhelpful (e.g., \emph{"I'm not sure what you mean."} or \emph{"Nothing happens."}). To actually provide models with meaningful feedback, we manually implement fine-grained errors such as \emph{"You can't move there as the door is closed."} or \emph{"You can't slice without a knife."} Ablation of the detailed prompt, finally reducing to the simple prompt, is shown in Figure~\ref{fig:prompt_sensitivity}.

The exact wordings of all prompts are shown below.

\subsection*{E.1: CoinCollector – Detailed \method Prompt (initial)}
\begin{lstlisting}[frame=single,style=appendixpy, numbers=none]
Please provide the output in strict JSON format, without any additional text or explanation, including a PDDL domain file as 'df' and a PDDL problem file as 'pf'. 
The format should strictly be:
    {
    "df": "...",
    "pf": "..."
    }

You are in an environment that you explore step by step. You must build and update PDDL files of the environment based on only your observations. 
Do not create something not appeared in the observations and also do not miss any observations e.g. through closed doors you may assume a room behind.
Do not assume that there will be a door connecting rooms.
Your task is always to keep exploration and go to a location you have not visited yet.
In other words, your goal should go to other not visited location.
If you enter a room, make sure you put everything you observed such as the direction in the problem file.

Here are your current observations: 
Action: look around
You are in the kitchen. To the North you see a closed plain door. To the East you see the corridor. 

Here are some valid actions you can take: ['close door to south', 'close door to west', 'move south', 'move west', 'open door to south', 'open door to west']

You should generate df and pf strictly follow this valid actions. There are in total 2 actions, that should exactly be the following two:
    1. :action open-door
        :parameters (?loc1 - location ?loc2 - location ?dir - direction)
    2. :action move
        :parameters (?from - location ?to - location ?dir - direction)
        
You should have a goal in the problem file like this: 
    (:goal 
        (at ?location)
    ) where location should be somewhere not visited
    
Note: in problem file's init, you shouldn't have "not ()" but only the single status
\end{lstlisting}

\subsection*{E.2: CoinCollector – Detailed \method Prompt (after generating \pf and \df)}
\begin{lstlisting}[frame=single,style=appendixpy, numbers=none]
Please provide the output in strict JSON format, without any additional text or explanation, including a PDDL domain file as 'df' and a PDDL problem file as 'pf'. 
The format should strictly be:
    {
    "df": "...",
    "pf": "..."
    }

You are in an environment that you explore step by step. You must build and update PDDL files of the environment based on only your observations. 
Do not create something not appeared in the observations and also do not miss any observations e.g. through closed doors you may assume a room behind.
Do not assume that there will be a door connecting rooms.
Your task is always to keep exploration and go to a location you have not visited yet.
In other words, your goal should go to other not visited location.
If you enter a room, make sure you put everything you observed such as the direction in the problem file.
Here are your current observations: Action: move east
You are in the corridor. To the West you see the kitchen. 

Here are some valid actions you can take: ['close door to south', 'close door to west', 'move south', 'move west', 'open door to south', 'open door to west']

You should generate df and pf strictly follow this valid actions. There are in total 2 actions, that should exactly be the following two:
1. :action open-door
    :parameters (?loc1 - location ?loc2 - location ?dir - direction)
2. :action move
    :parameters (?from - location ?to - location ?dir - direction)

You should have a goal in the problem file like this: 
(:goal 
    (at ?location)
) where location should be somewhere not visited

Note: in problem file's init, you shouldn't have "not ()" but only the single status

This is previous domain file: 
(define (domain exploration)
  (:requirements :strips)
  (:types location direction)
  (:predicates
    (at ?loc - location)
    (connected ?loc1 - location ?loc2 - location ?dir - direction)
    (door-closed ?loc1 - location ?loc2 - location ?dir - direction)
    (door-open ?loc1 - location ?loc2 - location ?dir - direction)
  )

  (:action open-door
    :parameters (?loc1 - location ?loc2 - location ?dir - direction)
    :precondition (door-closed ?loc1 ?loc2 ?dir)
    :effect (and (not (door-closed ?loc1 ?loc2 ?dir))
                 (door-open ?loc1 ?loc2 ?dir)
                 (connected ?loc1 ?loc2 ?dir))
  )

  (:action move
    :parameters (?from - location ?to - location ?dir - direction)
    :precondition (and (at ?from) (connected ?from ?to ?dir))
    :effect (and (not (at ?from))
                 (at ?to))
  )
)

This is previous problem file: (define (problem explore-environment)
  (:domain exploration)
  (:objects
    kitchen corridor room-north - location
    north east south west - direction
  )
  (:init
    (at kitchen)
    (door-closed kitchen room-north north)
    (connected kitchen corridor east)
  )
  (:goal (at corridor))
)

This is all the memory you have in this game including each action and its corresponding observations: 
Action: look around
You are in the kitchen. To the North you see a closed plain door. To the East you see the corridor. 
Action: move east
You are in the corridor. To the West you see the kitchen. 
    
Now modify those two files according to the new observations and notes. Fix any errors you made in the previous setting according to the new observation.
Generate updated files based on your new observation.
\end{lstlisting}

\subsection*{E.3: CoinCollector – Simple \method Prompt (initial)}
\begin{lstlisting}[frame=single,style=appendixpy, numbers=none]
Please provide the output in strict JSON format, without any additional text or explanation, including a PDDL domain file as 'df' and a PDDL problem file as 'pf'. 
The format should strictly be:
    {
    "df": "...",
    "pf": "..."
    }

You are in an environment that you explore step by step. You must build and update PDDL files of the environment based on only your observations. 
Do not create something not appeared in the observations and also do not miss any observations e.g. through closed doors you may assume a room behind.
Do not assume that there will be a door connecting rooms.
Your task is always to keep exploration and go to a location you have not visited yet.
In other words, your goal should go to other not visited location.
If you enter a room, make sure you put everything you observed such as the direction in the problem file.
Here are your current observations: Action: look around
You are in the kitchen. To the East you see a closed plain door. To the West you see the corridor. 

Here are some valid actions you can take: ['close door to south', 'close door to west', 'move south', 'move west', 'open door to south', 'open door to west']

You should generate df and pf strictly follow this valid actions. There are in total 2 actions, that should exactly be the following two:
1. :action open-door
    :parameters (?loc1 - location ?loc2 - location ?dir - direction)
2. :action move
    :parameters (?from - location ?to - location ?dir - direction)
    
Note: in problem file's init, you shouldn't have "not ()" but only the single status
\end{lstlisting}

\subsection*{E.4: CoinCollector – Simple \method Prompt (after generating \pf and \df)}
\begin{lstlisting}[frame=single,style=appendixpy, numbers=none]
Please provide the output in strict JSON format, without any additional text or explanation, including a PDDL domain file as 'df' and a PDDL problem file as 'pf'. 
The format should strictly be:
    {
    "df": "...",
    "pf": "..."
    }

You are in an environment that you explore step by step. You must build and update PDDL files of the environment based on only your observations. 
Do not create something not appeared in the observations and also do not miss any observations e.g. through closed doors you may assume a room behind.
Do not assume that there will be a door connecting rooms.
Your task is always to keep exploration and go to a location you have not visited yet.
In other words, your goal should go to other not visited location.
If you enter a room, make sure you put everything you observed such as the direction in the problem file.

Here are your current observations: 
Action: move north
You are in the corridor. To the North you see a closed wood door. To the South you see the kitchen. To the East you see a closed wood door. To the West you see a closed patio door. 

Here are some valid actions you can take: ['close door to south', 'close door to west', 'move south', 'move west', 'open door to south', 'open door to west']

You should generate df and pf strictly follow this valid actions. There are in total 2 actions, that should exactly be the following two:
1. :action open-door
    :parameters (?loc1 - location ?loc2 - location ?dir - direction)
2. :action move
    :parameters (?from - location ?to - location ?dir - direction)
    
Note: in problem file's init, you shouldn't have "not ()" but only the single status

This is previous domain file: 
(define (domain exploration)
  (:requirements :strips)
  (:types location direction)
  (:predicates
    (at ?loc - location)
    (connected ?loc1 - location ?loc2 - location ?dir - direction)
    (door ?loc1 - location ?loc2 - location ?dir - direction)
    (open ?loc1 - location ?loc2 - location ?dir - direction)
    (closed ?loc1 - location ?loc2 - location ?dir - direction)
  )

  (:action open-door
    :parameters (?loc1 - location ?loc2 - location ?dir - direction)
    :precondition (and (door ?loc1 ?loc2 ?dir) (closed ?loc1 ?loc2 ?dir))
    :effect (and (not (closed ?loc1 ?loc2 ?dir)) (open ?loc1 ?loc2 ?dir))
  )

  (:action move
    :parameters (?from - location ?to - location ?dir - direction)
    :precondition (and (at ?from) (connected ?from ?to ?dir)
                     (or (not (door ?from ?to ?dir)) (open ?from ?to ?dir)))
    :effect (and (not (at ?from)) (at ?to))
  )
)

This is previous problem file: (define (problem explore-kitchen)
  (:domain exploration)
  (:objects
    kitchen corridor room_s - location
    north south - direction
  )
  (:init
    (at kitchen)
    (connected kitchen corridor north)
    (connected corridor kitchen south)
    (door kitchen room_s south)
    (closed kitchen room_s south)
  )
  (:goal (at corridor))
)
This is all the memory you have in this game including each action and its corresponding observations: 
Action: look around
You are in the kitchen. To the North you see the corridor. To the South you see a closed frosted-glass door. 
Action: move north
You are in the corridor. To the North you see a closed wood door. To the South you see the kitchen. To the East you see a closed wood door. To the West you see a closed patio door. 

Now modify those two files according to the new observations and notes. Fix any errors you made in the previous setting according to the new observation.
Generate updated files based on your new observation.
\end{lstlisting}

\subsection*{E.5: CoinCollector – PlanGen Prompt}
\begin{lstlisting}[frame=single,style=appendixpy, numbers=none]
You are in an environment that you explore step by step. Based on your observations, generate a series of valid actions to progress in the environment.
Here are your current observations: {brief_obs}
Here are some valid actions you can take: {valid_actions}
Your goal is to explore new locations and interact with the environment effectively. Ensure actions are logical and do not repeat unnecessarily.

Additional context:
{overall_memory if overall_memory else "No additional memory available."}

If there are errors or obstacles, here is the message:
{large_loop_error_message if large_loop_error_message else "No errors or obstacles mentioned."}

Provide the output in strict JSON format like this, while you should only generate one action at a time:
{{
    "actions": ["action1"]
}}
\end{lstlisting}

\subsection*{E.6: ALFWorld – Detailed Prompt (initial)}
\begin{lstlisting}[frame=single,style=appendixpy, numbers=none]
Please provide the output in strict JSON format, without any additional text or explanation, including a PDDL domain file as 'df' and a PDDL problem file as 'pf'. 
The format should strictly be:
    {
    "df": "...",
    "pf": "..."
    }

You are in an environment that you must explore step by step. Your task is to build and update PDDL files for the environment using only your direct observations. Do not create or assume any objects, relationships, or details that have not been observed, and ensure you include all observations.

The environment is a room containing various objects. Some of these objects are on, in, or contained within other objects and receptacles. You will initially be located as init_receptacle. You can assume all receptacles are freely reachable.

Now, Your task is to: put some cloth on bathtubbasin.
Here are your current observations: Action: look around
You are in the middle of a room. Looking quickly around you, you see a bathtubbasin 1, a cabinet 5, a cabinet 4, a cabinet 3, a cabinet 2, a cabinet 1, a countertop 1, a garbagecan 1, a handtowelholder 2, a handtowelholder 1, a sinkbasin 1, a toilet 1, a toiletpaperhanger 1, and a towelholder 1.

Only the following actions are allowed: (There are only two types: object and receptacle)
    1. go to a receptacle
        :action GotoLocation
        :parameters (?from - receptacle ?to - receptacle)
    2. open a receptacle if it is closed
        :action OpenObject
        :parameters (?r - receptacle)
    3. close a receptacle
        :action CloseObject
        :parameters (?r - receptacle)
    4. take an object from another receptacle
        :action PickupObject
        :parameters (?o - object ?r - receptacle)
    5. put object into/on/in another receptacle
        :action PutObject
        :parameters (?o - object ?r - receptacle)
    6. using an object/receptacle by turning it on/off with a switch
        :action useObject
        :parameters (?o - object)
    7. heat an object using a receptacle
        :action HeatObject
        :parameters (?o - object ?r - microwaveReceptacle)
    8. clean an object using a receptacle
        :action CleanObject
        :parameters (?o - object ?r - sinkbasinReceptacle)
    9. cool an object using a receptacle
        :action CoolObject
        :parameters (?o - object ?r - fridgeReceptacle)
    10. slice an object using a sharp object
        :action SliceObject
        :parameters (?r - receptacle ?co - object ?sharp_o - sharpObject)

You must go to a receptacle first in order to use/open it or take/put objects from/on it.

The process involves two main stages:

1. Always searching for the aim Object first!!!
    In this stage, your goal is to go to and may need to open new, unvisited recepatacles until you find the object mentioned in the task. Some receptacles cannot be opened so you can directly see what objects after you go to that receptacle.

    You can only use the GotoLocation action to travel to a new location and the OpenObject action (if the receptacle is closed) to verify whether it contains the target object.

    Goal 1.1: Reach a location that has not been visited (the location should be a receptacle) using the GotoLocation action. 
        You goal should look like this:
        (:goal 
            (at ?recepatacle)
        ) where recepatacle should be somewhere or some recepatacles not visited.

    Goal 1.2: If you already go to the recepatacle and found the recepatacle is closed, use the OpenObject action to open it and inspect the contents. 
        Your goal should look like this:
        (:goal 
            (opened ?recepatacle)
        ) where recepatacle should be the recepatacle you want to open.

2. After you seeing the aim object in any receptacle, using the Object to Complete the Task:
    After you have located the object (the object may have some numbers added), you should always first pick up the object from that receptacle and update your goal to focus on how the object is used to complete the task. Remember your goal is Your task is to: put some cloth on bathtubbasin.. Based on different adjectives, you may need to perform different actions for the object in different ways.

    This may involve more than simply transferring it from one place to another.
    For example: You might examine the object or a nearby receptacle to gather information. You may need to use another tool or device (like a lamp or a switch). Some tasks require you to slice, heat, cool, or clean the object using an appropriate receptacle (e.g., microwave, sink, fridge).

    If necessary, use the PickupObject action to retrieve the item, and the GotoLocation action to move to the correct place.
    Then, apply the object in a purposeful way, not just move it, but interact with the environment to fulfill the task's actual goal.

    Hint: 
    1. If you want to heat, clean, and cool an object, after you go to that aim receptacle, do not put the object in the receptacle but do the action directly. For example, go to fridge, then cool the object with receptacle.
    2. If you want to slice an object, you should first go to the receptacle where both the sharp object and the aim object are located and ONLY pick up the sharp object then do the slice action. Don't forget to put the sharp object back to the receptacle after you finish slicing.
    3. If you want to examine or look at an object with a lamp, you should first go to the receptacle where the object is located and then pick it up and take the USE action of the lamp. You don't need to take the lamp but directly use it.
    4. If there are multiple actions needed to complete the task, you can break them down into smaller subgoals. For example, if you need to slice and then heat an object, first focus on slicing it, and then move on to heating it.

In summary, the first stage is all about finding the object. This might involve going to an unvisited receptacle and opening it if necessary.

Note: 
1. some receptacles have numbers in their names. Always keep them as they are. For example, "towelholder1" should not be changed to "towelholder".
2. Your initial goal should always be to go to a new location instead of put something into somewhere.
3. Do not enter stage 2 when not finishing stage 1.

Note: Always include :negative preconditions in your :requirements whenever you use (not) or delete effects, and never leave an :precondition or :effect block empty, either omit it or include at least one literal.

\end{lstlisting}

\subsection*{E.7: ALFWorld – Detailed Prompt (after generating \pf and \df)}
\begin{lstlisting}[frame=single,style=appendixpy, numbers=none]
Please provide the output in strict JSON format, without any additional text or explanation, including a PDDL domain file as 'df' and a PDDL problem file as 'pf'. 
The format should strictly be:
    {
    "df": "...",
    "pf": "..."
    }

You are in an environment that you must explore step by step. Your task is to build and update PDDL files for the environment using only your direct observations. Do not create or assume any objects, relationships, or details that have not been observed, and ensure you include all observations.

The environment is a room containing various objects. Some of these objects are on, in, or contained within other objects and receptacles. You will initially be located as init_receptacle. You can assume all receptacles are freely reachable.

Now, Your task is to: put some cloth on bathtubbasin.
Here are your current observations: 
Action: go to cabinet 1
You arrive at cabinet 1. The cabinet 1 is closed.

Only the following actions are allowed: (There are only two types: object and receptacle)
    1. go to a receptacle
        :action GotoLocation
        :parameters (?from - receptacle ?to - receptacle)
    2. open a receptacle if it is closed
        :action OpenObject
        :parameters (?r - receptacle)
    3. close a receptacle
        :action CloseObject
        :parameters (?r - receptacle)
    4. take an object from another receptacle
        :action PickupObject
        :parameters (?o - object ?r - receptacle)
    5. put object into/on/in another receptacle
        :action PutObject
        :parameters (?o - object ?r - receptacle)
    6. using an object/receptacle by turning it on/off with a switch
        :action useObject
        :parameters (?o - object)
    7. heat an object using a receptacle
        :action HeatObject
        :parameters (?o - object ?r - microwaveReceptacle)
    8. clean an object using a receptacle
        :action CleanObject
        :parameters (?o - object ?r - sinkbasinReceptacle)
    9. cool an object using a receptacle
        :action CoolObject
        :parameters (?o - object ?r - fridgeReceptacle)
    10. slice an object using a sharp object
        :action SliceObject
        :parameters (?r - receptacle ?co - object ?sharp_o - sharpObject)

You must go to a receptacle first in order to use/open it or take/put objects from/on it.

The process involves two main stages:

1. Always searching for the aim Object first!!!
    In this stage, your goal is to go to and may need to open new, unvisited recepatacles until you find the object mentioned in the task. Some receptacles cannot be opened so you can directly see what objects after you go to that receptacle.

    You can only use the GotoLocation action to travel to a new location and the OpenObject action (if the receptacle is closed) to verify whether it contains the target object.

    Goal 1.1: Reach a location that has not been visited (the location should be a receptacle) using the GotoLocation action. 
        You goal should look like this:
        (:goal 
            (at ?recepatacle)
        ) where recepatacle should be somewhere or some recepatacles not visited.

    Goal 1.2: If you already go to the recepatacle and found the recepatacle is closed, use the OpenObject action to open it and inspect the contents. 
        Your goal should look like this:
        (:goal 
            (opened ?recepatacle)
        ) where recepatacle should be the recepatacle you want to open.

2. After you seeing the aim object in any receptacle, using the Object to Complete the Task:
    After you have located the object (the object may have some numbers added), you should always first pick up the object from that receptacle and update your goal to focus on how the object is used to complete the task. Remember your goal is Your task is to: put some cloth on bathtubbasin.. Based on different adjectives, you may need to perform different actions for the object in different ways.

    This may involve more than simply transferring it from one place to another.
    For example: You might examine the object or a nearby receptacle to gather information. You may need to use another tool or device (like a lamp or a switch). Some tasks require you to slice, heat, cool, or clean the object using an appropriate receptacle (e.g., microwave, sink, fridge).

    If necessary, use the PickupObject action to retrieve the item, and the GotoLocation action to move to the correct place.
    Then, apply the object in a purposeful way, not just move it, but interact with the environment to fulfill the task's actual goal.

    Hint: 
    1. If you want to heat, clean, and cool an object, after you go to that aim receptacle, do not put the object in the receptacle but do the action directly. For example, go to fridge, then cool the object with receptacle.
    2. If you want to slice an object, you should first go to the receptacle where both the sharp object and the aim object are located and ONLY pick up the sharp object then do the slice action. Don't forget to put the sharp object back to the receptacle after you finish slicing.
    3. If you want to examine or look at an object with a lamp, you should first go to the receptacle where the object is located and then pick it up and take the USE action of the lamp. You don't need to take the lamp but directly use it.
    4. If there are multiple actions needed to complete the task, you can break them down into smaller subgoals. For example, if you need to slice and then heat an object, first focus on slicing it, and then move on to heating it.

In summary, the first stage is all about finding the object, this might involve going to an unvisited receptacle and opening it if necessary.

Note: 
1. some receptacles have numbers in their names. Always keep them as they are. For example, "towelholder1" should not be changed to "towelholder".
2. Your initial goal should always be to go to a new location instead of put something into somewhere.
3. Do not enter stage 2 when not finishing stage 1.

Note: Always include :negative preconditions in your :requirements whenever you use (not) or delete effects, and never leave an :precondition or :effect block empty, either omit it or include at least one literal.

This is previous domain file: 
(define (domain room_env)
  (:requirements :strips :typing :negative-preconditions)
  (:types receptacle microwaveReceptacle sinkbasinReceptacle fridgeReceptacle object sharpObject)
  (:predicates
    (at ?r - receptacle)
    (visited ?r - receptacle)
    (closed ?r - receptacle)
    (opened ?r - receptacle)
    (contains ?r - receptacle ?o - object)
    (holding ?o - object)
    (used ?o - object)
    (heated ?o - object)
    (cleaned ?o - object)
    (cooled ?o - object)
    (sliced ?o - object)
  )

  (:action GotoLocation
    :parameters (?from - receptacle ?to - receptacle)
    :precondition (and (at ?from) (not (visited ?to)))
    :effect (and (at ?to) (visited ?to) (not (at ?from)))
  )

  (:action OpenObject
    :parameters (?r - receptacle)
    :precondition (and (at ?r) (closed ?r))
    :effect (and (opened ?r) (not (closed ?r)))
  )

  (:action CloseObject
    :parameters (?r - receptacle)
    :precondition (and (at ?r) (opened ?r))
    :effect (and (closed ?r) (not (opened ?r)))
  )

  (:action PickupObject
    :parameters (?o - object ?r - receptacle)
    :precondition (and (at ?r) (contains ?r ?o))
    :effect (and (holding ?o) (not (contains ?r ?o)))
  )

  (:action PutObject
    :parameters (?o - object ?r - receptacle)
    :precondition (and (at ?r) (holding ?o))
    :effect (and (contains ?r ?o) (not (holding ?o)))
  )

  (:action useObject
    :parameters (?o - object)
    :precondition (and (at init_receptacle)) ; minimal precondition placeholder
    :effect (used ?o)
  )

  (:action HeatObject
    :parameters (?o - object ?r - microwaveReceptacle)
    :precondition (and (at ?r) (contains ?r ?o))
    :effect (heated ?o)
  )

  (:action CleanObject
    :parameters (?o - object ?r - sinkbasinReceptacle)
    :precondition (and (at ?r) (contains ?r ?o))
    :effect (cleaned ?o)
  )

  (:action CoolObject
    :parameters (?o - object ?r - fridgeReceptacle)
    :precondition (and (at ?r) (contains ?r ?o))
    :effect (cooled ?o)
  )

  (:action SliceObject
    :parameters (?r - receptacle ?co - object ?sharp_o - sharpObject)
    :precondition (and (at ?r) (contains ?r ?co) (contains ?r ?sharp_o))
    :effect (sliced ?co)
  )
)

This is previous problem file: 
(define (problem cloth_search)
  (:domain room_env)
  (:objects
    init_receptacle - receptacle
    bathtubbasin1 cabinet5 cabinet4 cabinet3 cabinet2 cabinet1 countertop1 garbagecan1 handtowelholder2 handtowelholder1 sinkbasin1 toilet1 toiletpaperhanger1 towelholder1 - receptacle
  )
  (:init
    (at init_receptacle)
    (visited init_receptacle)
    ;; Assume cabinet1 is closed and may contain the target cloth (even though cloth hasn't been observed yet), so we search there
    (closed cabinet1)
  )
  (:goal
    (at cabinet1)
  )
)
This is all the memory you have in this game including each action and its corresponding observations: 
Action: look around
You are in the middle of a room. Looking quickly around you, you see a bathtubbasin 1, a cabinet 5, a cabinet 4, a cabinet 3, a cabinet 2, a cabinet 1, a countertop 1, a garbagecan 1, a handtowelholder 2, a handtowelholder 1, a sinkbasin 1, a toilet 1, a toiletpaperhanger 1, and a towelholder 1.
Action: go to cabinet 1
You arrive at cabinet 1. The cabinet 1 is closed.

Now modify those two files according to the new observations and notes. Fix any errors you made in the previous setting according to the new observation.
Generate updated files based on your new observation.
\end{lstlisting}

\subsection*{E.8: ALFWorld – Simple Prompt (initial)}
\begin{lstlisting}[frame=single,style=appendixpy, numbers=none]
Please provide the output in strict JSON format, without any additional text or explanation, including a PDDL domain file as 'df' and a PDDL problem file as 'pf'. 
The format should strictly be:
    {
    "df": "...",
    "pf": "..."
    }

You are in an environment that you must explore step by step. Your task is to build and update PDDL files for the environment using only your direct observations. Do not create or assume any objects, relationships, or details that have not been observed, and ensure you include all observations.

The environment is a room containing various objects. Some of these objects are on, in, or contained within other objects and receptacles. You will initially be located as init_receptacle. You can assume all receptacles are freely reachable.

Now, Your task is to: put some cloth on bathtubbasin.
Here are your current observations: 
Action: look around
You are in the middle of a room. Looking quickly around you, you see a bathtubbasin 1, a cabinet 5, a cabinet 4, a cabinet 3, a cabinet 2, a cabinet 1, a countertop 1, a garbagecan 1, a handtowelholder 2, a handtowelholder 1, a sinkbasin 1, a toilet 1, a toiletpaperhanger 1, and a towelholder 1.


The following actions are allowed: (There are only two types: object and receptacle)
1. go to a receptacle
    :action GotoLocation
    :parameters (?from - receptacle ?to - receptacle)
2. open a receptacle if it is closed
    :action OpenObject
    :parameters (?r - receptacle)
3. close a receptacle
    :action CloseObject
    :parameters (?r - receptacle)
4. take an object from another receptacle
    :action PickupObject
    :parameters (?o - object ?r - receptacle)
5. put object into/on/in another receptacle
    :action PutObject
    :parameters (?o - object ?r - receptacle)
6. using an object/receptacle by turning it on/off with a switch
    :action useObject
    :parameters (?o - object)
7. heat an object using a receptacle
    :action HeatObject
    :parameters (?o - object ?r - microwaveReceptacle)
8. clean an object using a receptacle
    :action CleanObject
    :parameters (?o - object ?r - receptacle)
9. cool an object using a receptacle
    :action CoolObject
    :parameters (?o - object ?r - fridgeReceptacle)
10. slice an object using a sharp object
    :action SliceObject
    :parameters (?r - receptacle ?co - object ?sharp_o - object)

Your process involves two main stages with the following subgoals:

Stage 1: Search for the Target Object
    Goal 1.1: Move to a new, unvisited receptacle using the GotoLocation action.
    Goal 1.2: If the receptacle is closed, use the OpenObject action to reveal its contents.

Stage 2: Use the Object to Complete the Task
    Goal 2.1: Pick up the target object using the PickupObject action.
    Goal 2.2: Move to the appropriate location needed to fulfill the task.
    Goal 2.3: Interact with relevant objects or receptacles (e.g., heat, clean, cool, slice, or use) to accomplish the task.

In summary, the first stage is all about finding the object, this might involve going to an unvisited receptacle and opening it if necessary.

Note: 
1. some receptacles have numbers in their names. Always keep them as they are. For example, "towelholder1" should not be changed to "towelholder".
2. Your initial goal should always be to go to a new location instead of put something into somewhere.
3. Do not enter stage 2 when not finishing stage 1.

Note: Always include :negative preconditions in your :requirements whenever you use (not) or delete effects, and never leave an :precondition or :effect block empty, either omit it or include at least one literal.

\end{lstlisting}

\subsection*{E.9: ALFWorld – Simple Prompt (after generating \pf and \df)}
\begin{lstlisting}[frame=single,style=appendixpy, numbers=none]
Please provide the output in strict JSON format, without any additional text or explanation, including a PDDL domain file as 'df' and a PDDL problem file as 'pf'. 
The format should strictly be:
    {
    "df": "...",
    "pf": "..."
    }

You are in an environment that you must explore step by step. Your task is to build and update PDDL files for the environment using only your direct observations. Do not create or assume any objects, relationships, or details that have not been observed, and ensure you include all observations.

The environment is a room containing various objects. Some of these objects are on, in, or contained within other objects and receptacles. You will initially be located as init_receptacle. You can assume all receptacles are freely reachable.

Now, Your task is to: clean some lettuce and put it in countertop.
Here are your current observations: 
Action: go to cabinet 27
You arrive at cabinet 27. The cabinet 27 is closed.

Action: open cabinet 27
You open the cabinet 27. The cabinet 27 is open. In it, you see nothing.

Action: clean lettuce with cabinet 27
Nothing happens.


The following actions are allowed: (There are only two types: object and receptacle)
    1. go to a receptacle
        :action GotoLocation
        :parameters (?from - receptacle ?to - receptacle)
    2. open a receptacle if it is closed
        :action OpenObject
        :parameters (?r - receptacle)
    3. close a receptacle
        :action CloseObject
        :parameters (?r - receptacle)
    4. take an object from another receptacle
        :action PickupObject
        :parameters (?o - object ?r - receptacle)
    5. put object into/on/in another receptacle
        :action PutObject
        :parameters (?o - object ?r - receptacle)
    6. using an object/receptacle by turning it on/off with a switch
        :action useObject
        :parameters (?o - object)
    7. heat an object using a receptacle
        :action HeatObject
        :parameters (?o - object ?r - microwaveReceptacle)
    8. clean an object using a receptacle
        :action CleanObject
        :parameters (?o - object ?r - receptacle)
    9. cool an object using a receptacle
        :action CoolObject
        :parameters (?o - object ?r - fridgeReceptacle)
    10. slice an object using a sharp object
        :action SliceObject
        :parameters (?r - receptacle ?co - object ?sharp_o - object)

Your process involves two main stages with the following subgoals:

Stage 1: Search for the Target Object
    Goal 1.1: Move to a new, unvisited receptacle using the GotoLocation action.
    Goal 1.2: If the receptacle is closed, use the OpenObject action to reveal its contents.

Stage 2: Use the Object to Complete the Task
    Goal 2.1: Pick up the target object using the PickupObject action.
    Goal 2.2: Move to the appropriate location needed to fulfill the task.
    Goal 2.3: Interact with relevant objects or receptacles (e.g., heat, clean, cool, slice, or use) to accomplish the task.

In summary, the first stage is all about finding the object, this might involve going to an unvisited receptacle and opening it if necessary.

Note: 
1. some receptacles have numbers in their names. Always keep them as they are. For example, "towelholder1" should not be changed to "towelholder".
2. Your initial goal should always be to go to a new location instead of put something into somewhere.
3. Do not enter stage 2 when not finishing stage 1.

Note: Always include :negative preconditions in your :requirements whenever you use (not) or delete effects, and never leave an :precondition or :effect block empty, either omit it or include at least one literal.
    
This is previous domain file: 
(define (domain cleaning_task)
  (:requirements :strips :typing :negative-preconditions)
  (:types receptacle object)

  (:predicates
    (at ?r - receptacle)
    (closed ?r - receptacle)
    (open ?r - receptacle)
    (in ?o - object ?r - receptacle)
    (carry ?o - object)
    (used ?o - object)
    (clean ?o - object)
    (heated ?o - object)
    (cooled ?o - object)
    (sliced ?o - object)
  )

  (:action GotoLocation
    :parameters (?from - receptacle ?to - receptacle)
    :precondition (and (at ?from) (not (at ?to)))
    :effect (and (not (at ?from)) (at ?to))
  )

  (:action OpenObject
    :parameters (?r - receptacle)
    :precondition (and (at ?r) (closed ?r))
    :effect (and (open ?r) (not (closed ?r)))
  )

  (:action CloseObject
    :parameters (?r - receptacle)
    :precondition (and (at ?r) (open ?r))
    :effect (and (closed ?r) (not (open ?r)))
  )

  (:action PickupObject
    :parameters (?o - object ?r - receptacle)
    :precondition (and (at ?r) (in ?o ?r) (not (carry ?o)))
    :effect (and (carry ?o) (not (in ?o ?r)))
  )

  (:action PutObject
    :parameters (?o - object ?r - receptacle)
    :precondition (and (at ?r) (carry ?o) (open ?r))
    :effect (and (in ?o ?r) (not (carry ?o)))
  )

  (:action useObject
    :parameters (?o - object)
    :precondition (and (not (used ?o)))
    :effect (used ?o)
  )

  (:action HeatObject
    :parameters (?o - object ?r - receptacle)
    :precondition (and (at ?r) (in ?o ?r) (open ?r))
    :effect (heated ?o)
  )

  (:action CleanObject
    :parameters (?o - object ?r - receptacle)
    :precondition (and (at ?r) (in ?o ?r) (open ?r))
    :effect (and (clean ?o) (carry ?o) (not (in ?o ?r)))
  )

  (:action CoolObject
    :parameters (?o - object ?r - receptacle)
    :precondition (and (at ?r) (in ?o ?r) (open ?r))
    :effect (cooled ?o)
  )

  (:action SliceObject
    :parameters (?r - receptacle ?co - object ?sharp_o - object)
    :precondition (and (at ?r) (in ?co ?r) (not (sliced ?co)) (used ?sharp_o))
    :effect (sliced ?co)
  )
)

This is previous problem file: 
(define (problem cleaning_task_problem)
  (:domain cleaning_task)
  (:objects
    init_receptacle cabinet1 cabinet2 cabinet3 cabinet4 cabinet5 cabinet6 cabinet7 cabinet8 cabinet9 cabinet10
    cabinet11 cabinet12 cabinet13 cabinet14 cabinet15 cabinet16 cabinet17 cabinet18 cabinet19 cabinet20
    cabinet21 cabinet22 cabinet23 cabinet24 cabinet25 cabinet26 cabinet27
    coffeemachine1 countertop1 countertop2 diningtable1 drawer1 drawer2 drawer3 drawer4 drawer5 drawer6 drawer7 drawer8 drawer9 drawer10 drawer11 drawer12
    fridge1 garbagecan1 microwave1 sinkbasin1 stoveburner1 stoveburner2 stoveburner3 stoveburner4 toaster1 - receptacle
    lettuce - object
  )
  (:init
    (at init_receptacle)
    (open init_receptacle)
    (closed cabinet1) (closed cabinet2) (closed cabinet3) (closed cabinet4) (closed cabinet5)
    (closed cabinet6) (closed cabinet7) (closed cabinet8) (closed cabinet9) (closed cabinet10)
    (closed cabinet11) (closed cabinet12) (closed cabinet13) (closed cabinet14) (closed cabinet15)
    (closed cabinet16) (closed cabinet17) (closed cabinet18) (closed cabinet19) (closed cabinet20)
    (closed cabinet21) (closed cabinet22) (closed cabinet23) (closed cabinet24) (closed cabinet25)
    (closed cabinet26) (closed cabinet27)
    (closed coffeemachine1)
    (closed countertop1) (closed countertop2)
    (closed diningtable1)
    (closed drawer1) (closed drawer2) (closed drawer3) (closed drawer4) (closed drawer5)
    (closed drawer6) (closed drawer7) (closed drawer8) (closed drawer9) (closed drawer10)
    (closed drawer11) (closed drawer12)
    (closed fridge1)
    (closed garbagecan1)
    (closed microwave1)
    (closed sinkbasin1)
    (closed stoveburner1) (closed stoveburner2) (closed stoveburner3) (closed stoveburner4)
    (closed toaster1)
    (in lettuce cabinet27)
  )
  (:goal (and (clean lettuce) (in lettuce countertop1)))
)

This is all the memory you have in this game including each action and its corresponding observations: 
Action: look around
You are in the middle of a room. Looking quickly around you, you see a cabinet 27, a cabinet 26, a cabinet 25, a cabinet 24, a cabinet 23, a cabinet 22, a cabinet 21, a cabinet 20, a cabinet 19, a cabinet 18, a cabinet 17, a cabinet 16, a cabinet 15, a cabinet 14, a cabinet 13, a cabinet 12, a cabinet 11, a cabinet 10, a cabinet 9, a cabinet 8, a cabinet 7, a cabinet 6, a cabinet 5, a cabinet 4, a cabinet 3, a cabinet 2, a cabinet 1, a coffeemachine 1, a countertop 2, a countertop 1, a diningtable 1, a drawer 12, a drawer 11, a drawer 10, a drawer 9, a drawer 8, a drawer 7, a drawer 6, a drawer 5, a drawer 4, a drawer 3, a drawer 2, a drawer 1, a fridge 1, a garbagecan 1, a microwave 1, a sinkbasin 1, a stoveburner 4, a stoveburner 3, a stoveburner 2, a stoveburner 1, and a toaster 1.

    
Now modify those two files according to the new observations and notes. Fix any errors you made in the previous setting according to the new observation.
Generate updated files based on your new observation.

You have already generate files according to the observations. The df and pf can generate actions but after simulating,
it got those errors: 
In this step, you take the following actions and observations from those actions:
    Action: go to cabinet 27
    You arrive at cabinet 27. The cabinet 27 is closed.
    Action: open cabinet 27
    You open the cabinet 27. The cabinet 27 is open. In it, you see nothing.
    Action: clean lettuce with cabinet 27
    Nothing happens.
    Please review both files and fix them.

Now modify those two files according to the error message.
\end{lstlisting}

\subsection*{E.10: ALFWorld – PlanGen Prompt}
\begin{lstlisting}[frame=single,style=appendixpy, numbers=none]
You are in an environment that you explore step by step. Based on your observations, generate one valid action at a time to progress in the environment.
Your task is to interact with objects and receptacles to complete a goal step by step.

Your specific task goal: Your task is to: put a remotecontrol in armchair.

Here are your current observations: Action: look around
You are in the middle of a room. Looking quickly around you, you see a armchair 2, a armchair 1, a coffeetable 2, a coffeetable 1, a diningtable 1, a garbagecan 1, a sidetable 2, a sidetable 1, and a sofa 1.


Valid actions you can take (follow exactly this format, replacing the parts in brackets with actual object and location names. Do not include any brackets in your output):
    - go to [towelholder 1]
    - open [cabinet 2]
    - take [cloth 1] from [cabinet 3]
    - move [soap bar 1] to [sink basin 2]
    - use [desk lamp 1]
    - heat [bread 1] with [microwave 1]
    - clean [fork 1] with [sink basin 1]
    - cool [wine bottle 1] with [fridge 1]
    - slice [bread 1] with [knife 1]
Only replace the object and receptacle names (the words that were inside the brackets) with the actual values from the game environment. Do not change the structure. Do not include brackets.

Your actions should exactly follow this phrasing, do not invent new formats. Every action must correspond to a valid command from the environment.

You must go to a receptacle first in order to use/open it or take/put objects from/on it. You can assume all receptacles are freely reachable.

The process involves two main stages:

1. Always searching for the aim Object first!!!
    In this stage, your goal is to go to and may need to open new, unvisited recepatacles until you find the object mentioned in the task. Some receptacles cannot be opened so you can directly see what objects after you go to that receptacle.

    You can only use the GotoLocation action to travel to a new location and the OpenObject action (if the receptacle is closed) to verify whether it contains the target object.

2. Using the Object to Complete the Task:
    Once you have located and picked up the object, update your goal to focus on how the object is used to complete the task. This may involve more than simply transferring it from one place to another.
    For example: You might examine the object or a nearby receptacle to gather information. You may need to use another tool or device (like a lamp or a switch). Some tasks require you to slice, heat, cool, or clean the object using an appropriate receptacle (e.g., microwave, sink, fridge).

    If necessary, use the PickupObject action to retrieve the item, and the GotoLocation action to move to the correct place.
    Then, apply the object in a purposeful way, not just move it, but interact with the environment to fulfill the task's actual goal.

    Note: if you want to heat, clean, and cool an object, you can go to that receptacle then do the action directly without put the object into that receptacle.
        But if you want to slice an object, you should first go to the receptacle and pick up the sharp object then do the slice action.

In summary, the first stage is all about finding the object, this might involve going to an unvisited receptacle and opening it if necessary.

Note: 
1. some receptacles have numbers in their names. Always keep them as they are. For example, "towelholder1" should not be changed to "towelholder".
2. Your initial goal should always be to go to a new location instead of put something into somewhere.
3. Do not enter stage 2 when not finishing stage 1.

Memory of past steps:
Action: look around
You are in the middle of a room. Looking quickly around you, you see a armchair 2, a armchair 1, a coffeetable 2, a coffeetable 1, a diningtable 1, a garbagecan 1, a sidetable 2, a sidetable 1, and a sofa 1.


If there are errors or obstacles, here is the message:
No errors or obstacles mentioned.

Provide the output in strict JSON format like this while you should only generate one action at a time:
{
    "actions": ["action1"]
}
\end{lstlisting}

\section*{Appendix F: Licensing}
We use two publicly available simulated environments in our work. CoinCollector is licensed under the Apache License 2.0. ALFWorld is licensed under the MIT License. 

\section*{Appendix G: Use of AI Assistants}

AI Assistants are used for less than 10\% of coding and writing in this work.

\end{document}